%% file: iclr2025_conference.tex
\documentclass{article} 
\usepackage[T1]{fontenc}
\usepackage{iclr2025_conference,times}
\usepackage{graphicx}
\usepackage{amsmath}
\usepackage{amssymb}
\usepackage{booktabs}
\usepackage{enumitem}
\usepackage{algorithm}
\usepackage{algpseudocode}
\usepackage{bm}
\usepackage{multirow}
\usepackage{caption}
\usepackage{subcaption}
\usepackage{tikz}
\usepackage{tabularx}
\usepackage[most]{tcolorbox}
\usepackage{listings} 
\usepackage{xcolor}   
\usepackage{algorithm}
\usepackage{algpseudocode}
\usepackage{amsmath}
\usepackage{tcolorbox}
\usepackage{xcolor}
\usepackage{colortbl}
\usepackage{tabularx}
\tcbuselibrary{skins}

\usepackage{arydshln}

\usepackage{adjustbox}
\usepackage{natbib}
\usepackage{colortbl}

\usepackage{pifont}
\usepackage{fontawesome}
\usepackage{lipsum}
\usepackage{bbm}
\usepackage{dsfont}
\input{math_commands.tex}

\usepackage{pdfpages} 
\usepackage{hyperref}
\usepackage{url}
\usepackage{subcaption}
\usepackage{wrapfig}
\usepackage{siunitx}
\tcbuselibrary{skins}

\definecolor{CiteColor}{HTML}{9d595c}
\definecolor{BoxColor}{HTML}{f8f3eb}
\definecolor{Color1}{HTML}{faead3}
\definecolor{Color2}{HTML}{c1cbd7}

\definecolor{pastelGreen}{HTML}{C2DACF}
\definecolor{pastelBlue}{HTML}{C2DCE9}
\definecolor{pastelLavender}{HTML}{EAE1EF}
\definecolor{pastelCream}{HTML}{F8F3EB}
\definecolor{darkText}{RGB}{50, 50, 50} 

\newtcolorbox{algoboxblue}[2][]{
    enhanced,
    colback=pastelCream,            
    colframe=pastelBlue,            
    fonttitle=\bfseries,   
    coltitle=black,              
    colbacktitle=pastelBlue,        
    arc=3mm,                        
    title={#2},
    #1
}
\NewDocumentCommand{\yafu}
{ mO{} }{\textcolor{red}{\textsuperscript{\textit{yafu}}\textsf{\textbf{\small[#1]}}}}

\hypersetup{
    colorlinks=true,            
    linkcolor=CiteColor,              
    citecolor=CiteColor,             
    urlcolor=CiteColor,           
}

\lstdefinestyle{myjson}{
    language=json,
    basicstyle=\ttfamily\small, 
    breaklines=true,            
    breakatwhitespace=true,     
    postbreak=\mbox{\textcolor{red}{$\hookrightarrow$}\space}, 
    showstringspaces=false,
    stringstyle=\color{purple},
    keywordstyle=\color{blue},
    backgroundcolor=\color{blue!6!white} 
}
\iclrfinalcopy
\title{Towards an AI Musician: Synthesizing Sheet Music Problems for Musical Reasoning}

\author{Zhilin Wang\textsuperscript{1}\textsuperscript{2}\thanks{\quad Equal contributions. Work was done during Zhilin Wang's internship at Shanghai AI Laboratory.} \quad Zhe Yang\textsuperscript{3}\footnotemark[1]\quad
    \textbf{Yun Luo}\textsuperscript{2}\quad \textbf{Yafu Li}\textsuperscript{2}\footnotemark[2] \quad
    \textbf{Xiaoye Qu}\textsuperscript{2} \quad  
    \textbf{Ziqian Qiao}\textsuperscript{6}\quad \\
    \textbf{Haoran Zhang}\textsuperscript{4}\textsuperscript{2} \quad 
    \textbf{Runzhe Zhan\textsuperscript{5}\textsuperscript{2}} \quad \textbf{Derek F. Wong\textsuperscript{5}} \quad\textbf{Jizhe Zhou}\textsuperscript{\textbf{3}} \quad
    \textbf{Yu Cheng}\textsuperscript{\textbf{7}}\thanks{\quad Corresponding authors.}\\
    \textsuperscript{1} University of Science and Technology of China
    \textsuperscript{2} Shanghai AI Laboratory \\
    \textsuperscript{3} Sichuan University 
    \textsuperscript{4} Shanghai Jiao Tong University
    \textsuperscript{5} University of Macau \\
    \textsuperscript{6} Tsinghua University \textsuperscript{7} The Chinese University of Hong Kong \\
    \textbf{Contact:}
    \href{mailto:zhilin.nlp@gmail.com}{zhilin.nlp@gmail.com}, \href{mailto:yangzhe@stu.scu.edu.cn}{yangzhe@stu.scu.edu.cn}, \href{mailto:yafuly@gmail.com}{yafuly@gmail.com} \\
    \quad\quad\quad\quad\href{mailto:chengyu@cse.cuhk.edu.hk}{chengyu@cse.cuhk.edu.hk}
}
\begin{document}

\maketitle

\begin{abstract}

Enhancing the ability of Large Language Models (LLMs) and Multimodal Large Language Models (MLLMs) to interpret sheet music is a crucial step toward building AI musicians. 
However, current research lacks both evaluation benchmarks and training data for sheet music reasoning. 
{Inspired by mathematics, where simple operations yield infinite verifiable problems, we introduce a novel approach that treats core music theory rules, such as those governing beats and intervals, as programmatic functions to systematically synthesize a vast and diverse corpus of sheet music reasoning problems.} 
This approach allows us to introduce a data synthesis framework that generates verifiable sheet music questions in both textual and visual modalities, leading to the Synthetic Sheet Music Reasoning Benchmark (SSMR-Bench) and a complementary training set. 
Evaluation results on SSMR-Bench highlight the key role reasoning plays in interpreting sheet music, while also pointing out the ongoing challenges in understanding sheet music in a visual format.
By leveraging synthetic data for RLVR, all models show significant improvements on the SSMR-Bench. Additionally, they also demonstrate considerable advancements on previously established human-crafted benchmarks, such as MusicTheoryBench and the music subset of MMMU.
Finally, our results show that the enhanced reasoning ability can also facilitate music composition. 

\end{abstract}

\input{article/intro}

\input{article/framework}
\input{article/experiment}

\input{article/related_work}

\input{article/conclusion}

\section*{ETHICS STATEMENT}

We believe there are no major ethical concerns associated with our work. The study is exclusively focused on the music domain, specifically sheet music reasoning, and does not involve any applications that could be deemed sensitive or harmful.

\section*{Reproducibility statement}

We make our code and the generated data available at \href{https://anonymous.4open.science/r/temp-179B}{Anonymous GitHub \faGithub}. This repository includes our framework for synthesizing sheet music reasoning problems, along with the corresponding evaluation scripts.
Besides, we provide the generated data, covering both textual and visual modalities.
\bibliography{iclr2025_conference}
\bibliographystyle{iclr2025_conference}

\appendix
\input{article/appendix}

\end{document}

%% file: math_commands.tex
\usepackage{amsmath,amsfonts,bm}

\def\eqref#1{equation~\ref{#1}}

\def\1{\bm{1}}

\DeclareMathAlphabet{\mathsfit}{\encodingdefault}{\sfdefault}{m}{sl}
\SetMathAlphabet{\mathsfit}{bold}{\encodingdefault}{\sfdefault}{bx}{n}



%% file: article/intro.tex
\section{Introduction}

\begin{quote}
\textit{"The sheet music is the language of musicians."}
\end{quote}



\begin{wrapfigure}{r}{0.5\textwidth}
    \vspace{-12mm}
    \centering
    \includegraphics[width=1\linewidth]{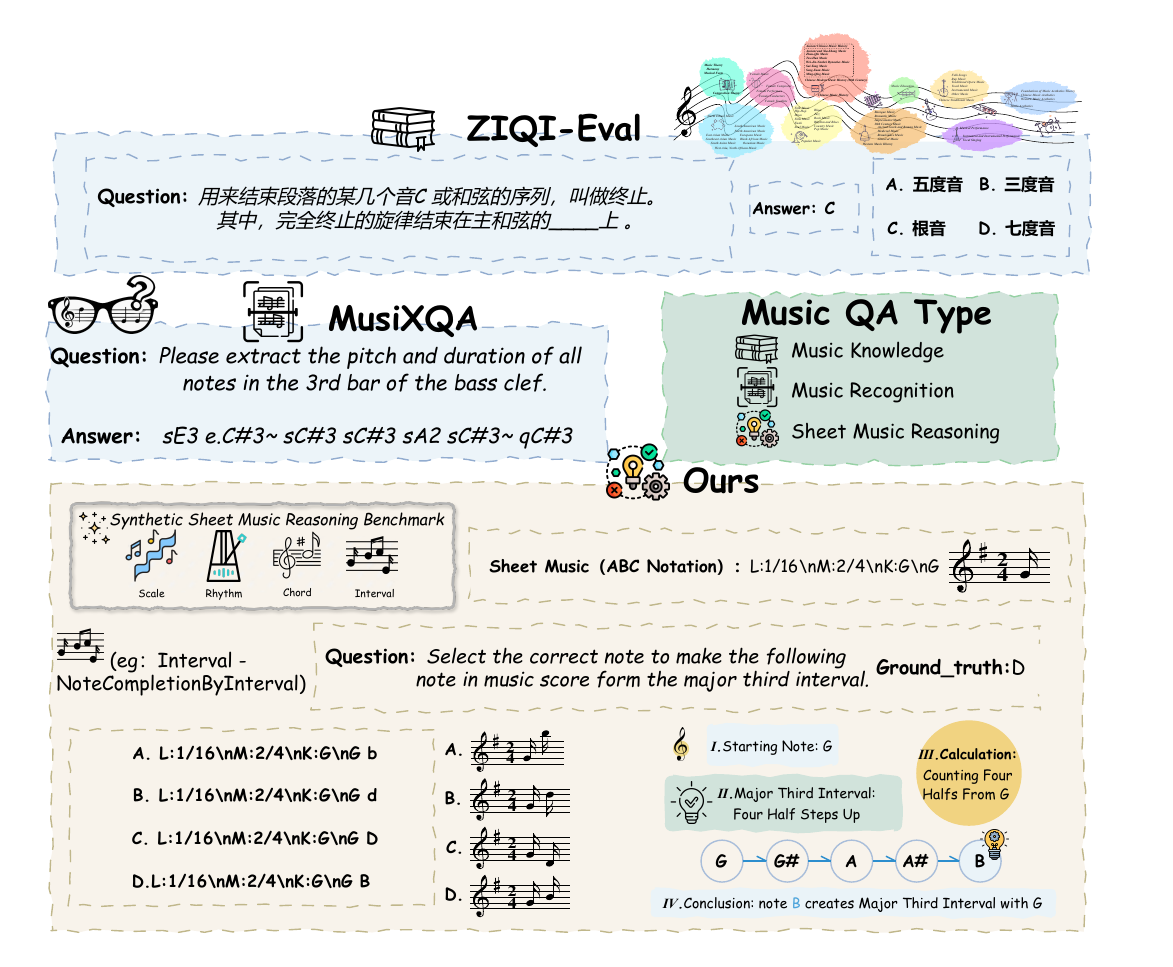}
    \caption{Comparison of Sheet Music Reasoning QA, Knowledge QA, and Music Recognition.}
    \label{fig:intro}
    \vspace{-4mm}
\end{wrapfigure}

Recent advancements in Large Language Models (LLMs) and Multimodal Large Language Models (MLLMs) have inspired researchers to explore the potential of developing AI musicians~\citep{qu_mupt_2025,bradshaw2025ariamididatasetpianomidi,wang2024whole}. Given that sheet music is the universal language of musicians, the ability to read and interpret it is an essential step for AI musicians~\citep{yuan2024chatmusician,wang_notagen_2025}.
We term this capability sheet music reasoning.
As illustrated in Figure~\ref{fig:intro}, sheet music reasoning differs fundamentally from Music Knowledge QA~\citep{li2024musicmaestromusicallychallenged}, which evaluates memorized knowledge, and from sheet music recognition~\citep{chen_musixqa_2025}, which focuses on identifying notation from images. 
Instead, sheet music reasoning requires applying learned musical knowledge to actively interpret and analyze sheet music. This task requires not only accurate recognition of musical symbols but also a nuanced grasp of their interactions within broader musical structures.




However, research on sheet music reasoning is still limited, with few standardized benchmarks. For instance, ChatMusician~\citep{yuan2024chatmusician} introduces MusicTheoryBench, 367 human-designed questions partly targeting sheet music reasoning in ABC notation. MMMU~\citep{yue2024mmmu} evaluates MLLMs on large-scale, college-level tasks, including some sheet music reasoning in image format. Yet both datasets are fully human-crafted, limiting scalability for training.



Inspired by mathematics, where simple operations yield limitless verifiable problems, we propose to leverage the rules of music theory to programmatically generate an endless supply of verifiable reasoning problems about sheet music. Such a resource enables robust evaluation and supports training with verifiable rewards~\citep{deepseekai2025deepseekr1incentivizingreasoningcapability} to improve musical reasoning. We introduce a novel data synthesis framework that produces questions in both textual and visual staff notation, organized into nine templates across Rhythm, Chord, Interval, and Scale. Using this framework, we construct the Synthetic Sheet Music Reasoning Benchmark (SSMR-Bench), comprising 1,600 textual and 1,600 visual QA pairs for evaluation, plus 8,000 pairs per modality for training.



We evaluate SSMR-Bench using a range of state-of-the-art LLMs and MLLMs. For the textual QA task, which represents sheet music in ABC notation, Qwen3-32B demonstrates performance on par with Deepseek-V3, while Deepseek-R1 exceeds Deepseek-V3 by 18.69\% overall. This highlights the model's enhanced reasoning capability in understanding sheet music.
For the visual QA task, Gemini2.5-Pro achieves the best performance with an average score of 56.62\%, surpassing Qwen2.5-VL-72B-Instruct, InternVL3-78B, and other models.
However, this performance is considerably lower than Gemini2.5-Pro’s results on the same questions presented in textual format, highlighting the challenges of interpreting sheet music from images.

We train a range of models on our synthetic dataset using Group Relative Policy Optimization (GRPO, ~\citealp{shao2024deepseekmath}). Across different model families (Qwen,~\citealp{yang_qwen3_2025} and Llama,~\citealp{grattafiori2024llama3herdmodels}), sizes (3B–8B), and modalities (text and image), all models achieve substantial improvements on SSMR-Bench. Notably, Qwen3-8B-Base improves by 13.06\%, reaching performance comparable to GPT-4.1-mini, while Qwen2.5-VL-7B-Instruct exhibits stronger multimodal reasoning, surpassing Gemini2.5-Pro.
Beyond the in-distribution benchmark, evaluation on external benchmarks such as MusicTheoryBench~\citep{yuan2024chatmusician} and MMMU Music~\citep{yue2024mmmu}, also demonstrates consistent improvement across all models. Interestingly, training on synthetic musical data also improves mathematical reasoning, underscoring the generality of the elicited capabilities.
Moreover, our evaluations show that the enhanced reasoning ability leads to more rhythmically precise sheet music composition, with the trained Qwen3-8B-Base outperforming Qwen3-8B-Thinking and demonstrating greater musical coherence and accuracy.

In conclusion, we summarize our key contributions as follows:

\begin{itemize}
    \item 
    We are the first to leverage music theory rules to programmatically synthesize verifiable sheet music problems, using them as both an evaluation benchmark and a training set for Reinforcement Learning with Verifiable Rewards (RLVR).
    
    \item We develop a data synthesis framework based on this idea, which is capable of generating verifiable sheet music questions with staff notation in both textual and visual modalities, and we provide SSMR-Bench along with a corresponding training set in both modalities.
    
    \item We conduct comprehensive experiments to demonstrate the potential of automatically synthesizing verifiable sheet music problems based on music theory rules. Training on this synthetic data enhances models’ reasoning abilities in sheet music, and this improved ability also shows promise in facilitating music composition.
\end{itemize}

%% file: article/framework.tex
\section{Synthesize Music Reasoning Questions}

\subsection{Fundamentals of Sheet Music}

Music is traditionally written on a staff, a set of five lines where visual symbols represent the core elements of music theory. The vertical position of a notehead indicates its pitch, while its shape (e.g., hollow or filled, with a stem or flag) defines its rhythm, or duration. These visual cues are organized into measures, separated by barlines, according to a time signature that dictates the underlying rhythmic pulse. 
While staff notation is intuitive for human musicians, its graphical nature is cumbersome for programmatic analysis. To bridge this gap, we utilize ABC notation, a text-based format that encodes the same musical information in a machine-readable way~\citep{qu_mupt_2025}. In ABC, pitches are represented by letters (C, D, E), rhythmic values are specified with numerical modifiers, and measures are delineated by vertical bars (|). This format allows us to translate the abstract rules of music theory into parsable strings. Examples of ABC notation are provided in Appendix~\ref{app:data}.

\subsection{Data Synthesis Framework}
\begin{figure}[t]
    \centering
      \includegraphics[width=\linewidth]{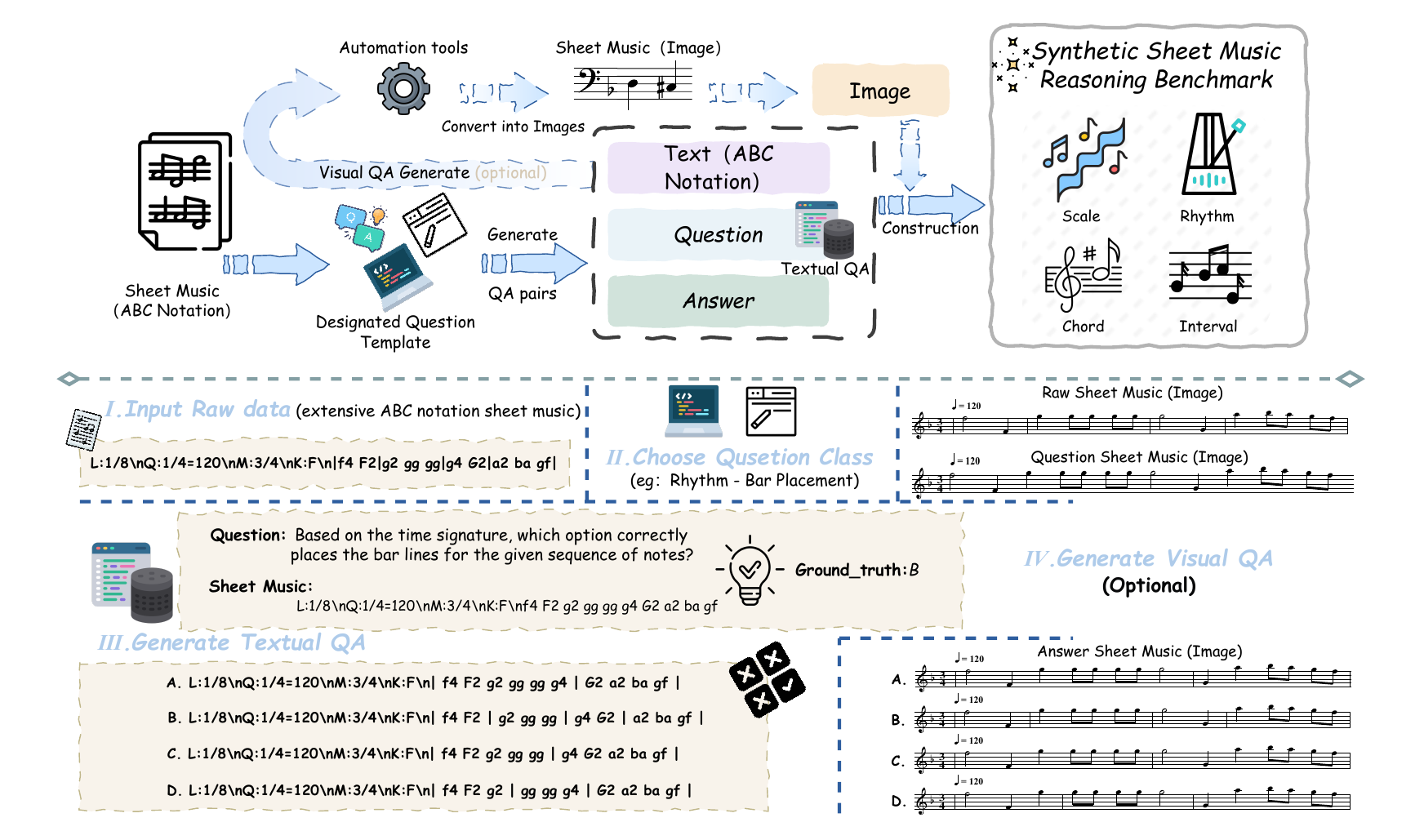}
    \caption{
    Overview of synthesizing verifiable sheet music questions. The upper plot illustrates the pipeline. Experts define rule-based Question Template Classes that generate questions and answers from existing sheet music in ABC notation, resulting in textual QA problems. The sheet music is then converted into an image, allowing the creation of Visual Question Answering (VQA)-style questions.
    The process is fully programmatic, without LLMs.
    The lower plot presents examples of generated textual and visual QA tasks. 
    }
    \label{fig:framework}
    \vspace{-10pt}
\end{figure}

\begin{wrapfigure}{r}{0.56\columnwidth} 
    \vspace{-5mm}
    \begin{algoboxblue}{Example: Bar Placement Question Synthesis}
        \label{alg:bar_placement_wrap}
        \begin{algorithmic}[1] 
            \Require $\mathcal{S}_{\text{orig}}$ (ABC notation)
            \Function{GenerateQA}{$\mathcal{S}_{\text{orig}}$}
                \State $\mathcal{A}_{\text{corr}} \gets \mathcal{S}_{\text{orig}}$
                \State $\mathcal{C}_{\text{txt}} \gets \Call{RemoveBars}{\mathcal{S}_{\text{orig}}}$
                \State $\mathcal{A}_{\text{incorr}} \gets \Call{GenDistractors}{\mathcal{C}_{\text{txt}}}$
                \State $\text{Choices} \gets \Call{Shuffle}{\{\mathcal{A}_{\text{corr}}\} \cup \mathcal{A}_{\text{incorr}}}$
                \State $\mathcal{Q}_{\text{txt}} \gets (\mathcal{C}_{\text{txt}}, \text{Choices})$
                \State $\mathcal{Q}_{\text{vis}} \gets \Call{RenderImgs}{\mathcal{Q}_{\text{txt}}}$
                \State \Return $(\mathcal{Q}_{\text{txt}}, \mathcal{Q}_{\text{vis}})$
            \EndFunction
        \end{algorithmic}
    \end{algoboxblue}
    \caption{Algorithm for bar placement question synthesis, including bar removal, distractor generation, option shuffling, and rendering of textual and visual questions.}
    \label{fig:qa_synthesis_algrithm}
    \vspace{-3mm}
\end{wrapfigure}

Our framework is designed to generate questions that test a model's understanding of the fundamental music principles, from rhythmic calculation and chord identification to the analysis of intervals and scales.
We outline the process of synthesizing sheet music questions based on music theory rules in Figure~\ref{fig:framework}. 
Sheet music is first obtained in ABC notation. Then, the experts design specialized Question Template Classes derived from music theory rules, each capable of generating both questions and corresponding answers from the ABC representation. Applying these templates produces textual QA pairs. Subsequently, the sheet music in each question can be converted into images using \textit{abcm2ps}
and \textit{ImageMagick}
, enabling the construction of VQA questions. The entire pipeline is fully programmatic, without reliance on LLMs, ensuring controllability and interpretability.

The lower plot in Figure~\ref{fig:framework} presents a visual illustration of an example generated by the Bar Placement Class. In this case, the bar lines of the original sheet music are removed, and the resulting incomplete score serves as input. According to music theory, the missing bar lines can be reconstructed by reasoning from the time signature, requiring models to infer rhythmic grouping and meter to restore the bar-line structure. Once the textual QA question is obtained, the sheet music in textual modality can be converted into images to generate the corresponding VQA question. Figure~\ref{fig:qa_synthesis_algrithm} provides the pseudocode for the Bar Placement Class, demonstrating how it functions.

\subsection{Data Construction}

\begin{table*}[!h]
    \centering    
    \small
    \caption{Comparison of music-related QA datasets across different aspects, where checkmarks (\ding{51}) indicate presence and crosses (\ding{55}) indicate absence.}
    \begin{tabular}{@{}l c c c c c c @{}}
        \toprule
        \multirow{2}{*}{\textbf{Dataset}} & \multirow{2}{*}{\textbf{Sheet Music QA}} & \multirow{2}{*}{\textbf{Synthetic}} & \multicolumn{2}{c}{\textbf{Modality}} & \multirow{2}{*}{\textbf{Trainable}} \\
        \cmidrule(lr){4-5}
         & & & \textbf{Textual} &  \textbf{Visual} &         \\
        \midrule
        MMMU~\citep{yue2024mmmu}             & \ding{51} & \ding{55} & \ding{55}   &\ding{51}      & \ding{55}    \\
        MusiXQA~\citep{chen_musixqa_2025}          & \ding{55}       & \ding{51} &\ding{55}   &\ding{51}  & \ding{51}    \\
        ZIQI-Eval~\citep{li2024musicmaestromusicallychallenged}     & \ding{55}       & \ding{55} &\ding{51}   &\ding{55}  & \ding{55}    \\
        MusicTheoryBench~\citep{yuan2024chatmusician} & \ding{51} & \ding{55} & \ding{51}   &\ding{55}    & \ding{55}  \\
        SSMR-Bench (Ours)           & \ding{51}       & \ding{51} & \ding{51}   &\ding{51}  & \ding{51} \\
        \bottomrule
    \end{tabular}
    \label{tab:dataset-comparison}
\end{table*}

The scalable synthesis framework we designed currently provides 9 types of questions. 
All questions are categorized into four categories: Rhythm, Chord, Interval, and Scale, which are the fundamental elements of sheet music.
Detailed descriptions of the question types and categories can be found in Appendix~\ref{app:data}.
To solve these questions, LLMs and MLLMs need basic music knowledge, sheet music reading skills, and reasoning ability.

\begin{wrapfigure}{r}{0.5\textwidth}
    \vspace{-4mm}
    \centering
    \includegraphics[width=0.8\linewidth]{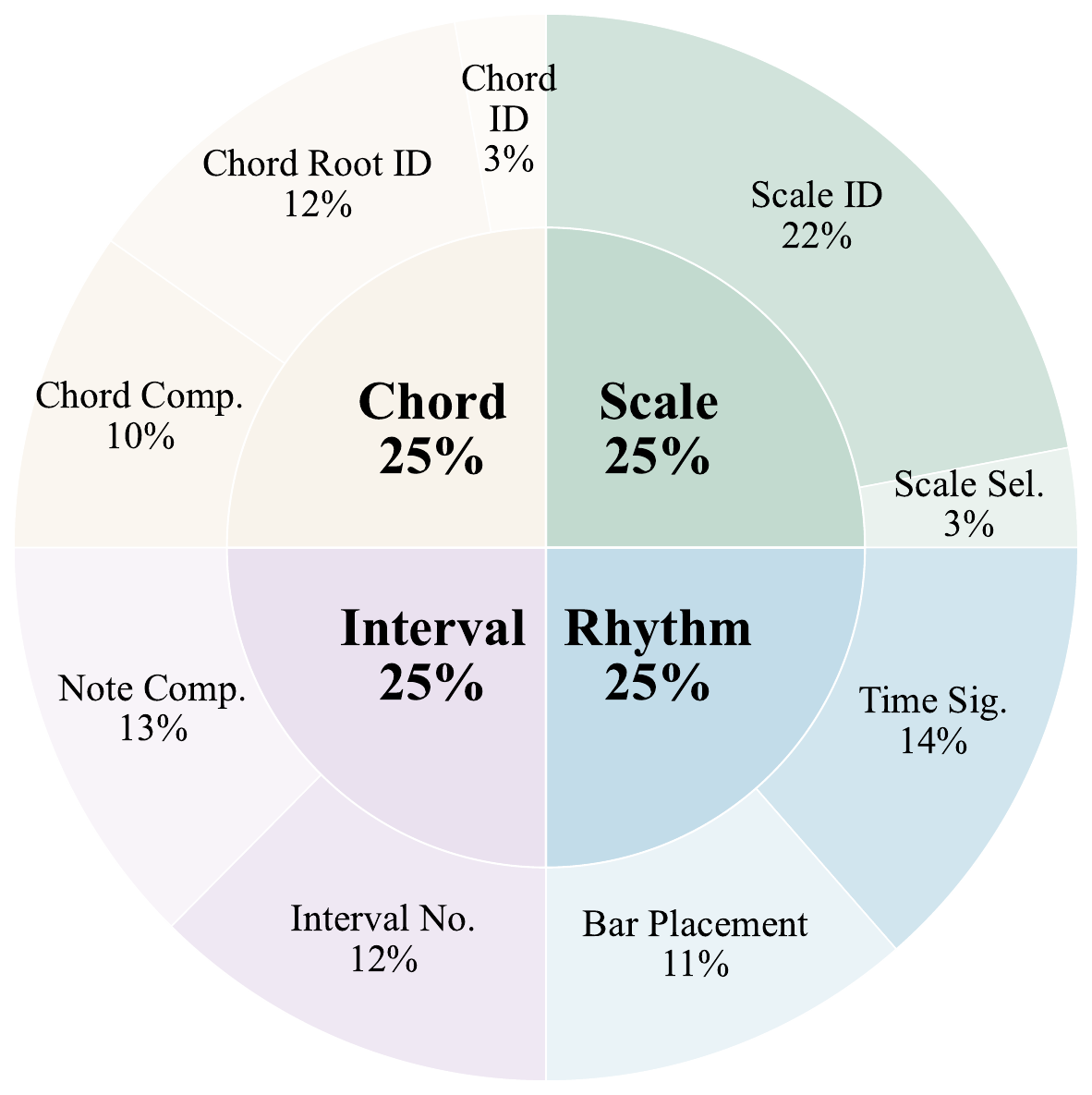}
    \caption{The composition of the SSMR-Bench test set consists of four categories: Rhythm, Chord, Interval, and Scale. Each category contributes equally, accounting for 25\% of the total.  
    }
    \label{fig:benchmark}
    \vspace{-8mm}
\end{wrapfigure}

Through our data synthesis framework, we introduce the Synthetic Sheet Music Reasoning Benchmark (SSMR-Bench), comprising 1,600 textual and 1,600 visual questions. 
The sheet music used to generate questions is sampled from MelodyHub~\citep{wu_melodyt5_2024}, a large-scale dataset in ABC notation.
The distribution of question types and categories within the benchmark is illustrated in Figure~\ref{fig:benchmark}. 
Furthermore, we offer a comprehensive training set, consisting of 8,000 textual and 8,000 visual questions. A comparative analysis between our framework and prior related benchmarks is presented in Table~\ref{tab:dataset-comparison}. 
Our framework produces both visual and textual modality questions for sheet music problems and constructs a verifiable training set without human curation, thereby enabling scalable data generation.

\subsection{Verifiable Music Training}

Following the RLVR paradigm, we employ the Group Relative Policy Optimization (GPRO,~\citealp{shao2024deepseekmath}) algorithm for RL training. GPRO approximates the advantage of a trajectory by normalizing its reward against the mean and standard deviation of rewards from $N$ sampled solutions for a given query:
\begin{gather}
    A_i = \frac{R(\tau_i) - \text{mean}(\{R(\tau_j) \mid j=1, \ldots, N\})}{\text{std}(\{R(\tau_j) \mid j=1, \ldots, N\})},
\end{gather}
where $R(\tau_i)$ is the reward for trajectory $\tau_i$. The rewards are calculated over $N$ trajectories sampled from the previous policy $\pi_{\theta_{\text{old}}}$. This relative scoring helps the model learn which outputs perform better or worse than the group's average. Then, the GPRO objective is given by:
\begin{equation}
\begin{gathered}
\mathcal{J}_{\text{GRPO}}(\theta) = \frac{1}{\sum_{i=1}^N |\tau_i|} \sum_{i=1}^N \sum_{t=1}^{|\tau_i|} \text{CLIP}(r_{i,t}(\theta), A_i, \epsilon) - \beta \cdot \mathbb{D}_{\text{KL}}[\pi_\theta \| \pi_{\text{ref}}].
\end{gathered}
\label{eq:grpo}
\end{equation}
Here, ${r_{i,t}(\theta) = {\pi_\theta(\tau_{i,t} | q, \tau_{i,<t})}/{\pi_{\theta_{\text{old}}}(\tau_{i,t} | q, \tau_{i,<t})}}$ is the importance sampling ratio, adjusting for the fact that the trajectories were generated by the old policy $\pi_{\theta_{\text{old}}}$, ensuring unbiased policy updates.

%% file: article/experiment.tex
\section{Experiment}

\subsection{Experimental Setup}

\paragraph{Evaluation.} We select the frontier LLMs and MLLMs to evaluate their performance on our SSMR-bench. For testing, the temperature is set to 0.7, with a length limit of 8192. We shuffle the multiple-choice options to avoid contamination. We utilize Math-Verify to parse the answers from the model responses and vLLM as the backend for inference serving. 


\paragraph{Implementation Details.} We employ multiple LLMs and MLLMs for reinforcement learning on our synthetic data. For the textual modality, we choose Qwen3-4B-Base, Qwen3-8B-Base, and Llama3.1-8B-IT (with “IT” referring to “Instruct”). For the visual modality, we select Qwen2.5-VL-7B-Instruct. Additionally, we include a variant of Qwen3-8B-Base trained on 8,000 math problems sampled from OpenR1~\citep{openr1} to compare the effectiveness of synthetic music reasoning data. Training is conducted using the GRPO algorithm with a batch size of 128, a rollout number of 8, and an update batch size of 64 over 300 training steps. The KL loss and entropy loss are removed. The reward function is based solely on accuracy, without incorporating format-related signals. The training template is provided in Appendix~\ref{app:training-template}.

\subsection{Results on Synthetic Music Reasoning Benchmark}
\label{sec:results-on-SSMR-bench}

\begin{table*}[!t]
    \setlength{\tabcolsep}{10pt}
    \centering
    \caption{Evaluation results of frontier models for textual and visual QA on the SSMR-bench. The table presents the performance of various models across different sheet music reasoning tasks, including Rhythm, Chord, Interval, and Scale, with the overall performance summarized for both textual and visual QA. “IT” is an abbreviation for “Instruct”.
    }
    \small
    \begin{tabular}{@{}l c c c c c >{\columncolor{Color1!40}}c}
        \toprule
        \multirow{3.5}{*}{\centering\textbf{Models}} &  \multirow{3.5}{*}{\centering\textbf{Thinking}} & \multicolumn{5}{c}{\textbf{\textit{Synthetic Sheet Music Reasoning Benchmark}}} \\
        \cmidrule(lr){3-7}
        &&\textbf{Rhythm} & \textbf{Chord} & \textbf{Interval} & \textbf{Scale} & \textbf{Overall} \\
        & & $(400)$ & $(400)$ & $(400)$ & $(400)$ & $(1600)$ \\
    
        \midrule
      
        \rowcolor[HTML]{C2DCE9}
        \multicolumn{7}{c}{\textbf{\textit{Textual QA}}} \\
        \midrule
        GPT-5 & \ding{51} & $97.75$ & $97.50$ & $92.75$ &	$100.00$ & $97.00$ \\	
        Gemini-2.5-Pro  & \ding{51} & $98.25$ & $96.50$ & $90.25$ &	$99.75$ & $96.19$ \\
        Deepseek-R1  & \ding{51}& $93.00$ 	& $90.50$& $91.25$&	$99.75$&$93.63$ \\
        GPT-4.1  & \ding{55} & $88.75$ & $68.75$ & $63.75$ & $98.00$ & $79.81$ \\
        Qwen3-32B-Thinking  &\ding{51} & $80.00$ & $75.00$ & $62.25$ & $89.50$ & $76.69$ \\        
        Deepseek-V3-250324  & \ding{55}& $78.75$ & $71.75$ & $58.75$ & $90.50$ & $74.94$ \\
        QwQ-32B  &\ding{51} & $69.75$ & $72.50$ & $59.50$ & $87.75$ & $72.38$ \\
        GPT-4.1-mini  & \ding{55}& $77.00$ &  $69.75$ &	 $56.25$ &  $83.75$ &	$71.69$ \\
        Qwen3-8B-Thinking       & \ding{51} & $54.00$ & $60.50$ & $50.75$ & $66.25$ & $57.88 $\\
        Qwen2.5-32B-IT  & \ding{55}& $54.50$ & $47.00$ & $53.75$ & $49.75$ & $51.25$ \\
        GPT-4o-mini  &\ding{55} & $48.00$ &	$54.50$	& $48.50$	& $40.25$	& $47.81$ \\
        Llama3.1-70B-IT  & \ding{55}& $39.75$ & $41.00$ & $47.00$ & $48.00$ & $43.94$ \\
        Llama3.1-8B-IT  & \ding{55} &	$23.50$ &	$19.00$ & $18.00$ & $23.25$ & $20.94$ \\
        
        \midrule
        \rowcolor[HTML]{C2DCE9}
        \multicolumn{7}{c}{\textbf{\textit{Visual QA}}} \\
        \midrule
        Gemini-2.5-Pro  &\ding{51} & $44.75$ &	$61.25$ &	 $32.25$	& $83.50$	& $55.44$ \\
        InternVL3-78B  &\ding{55} & $34.25$ & $44.75$ & $29.00$ & $75.00$ & $45.75$ \\
        Qwen2.5-VL-72B-IT  & \ding{55}& $39.75$ & $41.75$ & $27.50$ & $73.25$ & $45.56$ \\
        InternVL2.5-78B  &\ding{55} & $29.00$ & $31.00$ & $32.25$ & $69.50$ & $40.44$ \\
       \bottomrule
    \end{tabular}
    \label{tab:main-results}
\end{table*}

Table~\ref{tab:main-results} presents the performance of advanced models on SSMR-bench, covering both LLMs and MLLMs in the textual and visual modalities. The performance metrics reveal notable variations between different models, highlighting their strengths and limitations in each modality.

\paragraph{Textual Modality Performance Comparison.}
The results for textual QA clearly show that LLMs, such as GPT-5, outperform other models across all categories. Specifically, GPT-5 attains the highest overall accuracy at 97.00\%, followed by Gemini-2.5-Pro at 96.10\%, DeepSeek-R1 at 93.63\%, and GPT-4.1 at 79.81\%.
In contrast, models like Qwen2.5-32B-IT and Llama3.1-70B-IT exhibit lower overall performance, with scores of 51.25\% and 43.94\%, respectively. Llama3.1-8B-IT shows the lowest performance, with an overall score of 20.94\%.

\paragraph{Performance of MLLMs on Visual Modality.}

The performance of MLLMs on sheet music reasoning, with staff notation presented in image format, is substantially lower than that of LLMs on textual reasoning. The best-performing MLLM is Gemini 2.5 Pro, which is recognized as one of the most advanced models available. It achieves an overall accuracy of 55.44\%, with particularly strong performance on Scale (84.75\%). Other models, such as Qwen2.5-VL-72B-Instruct, InternVL3-78B, and InternVL2.5-78B, show accuracies of 45.56\%, 45.75\%, and 40.44\%, respectively. 

\paragraph{Cross-Modality Performance Analysis.}

The lower performance of MLLMs in the visual modality can be attributed to the unique challenges posed by music element recognition. 
For Gemini-2.5-Pro, accuracy reaches 96.19\% when sheet music is presented in textual modality, compared to 55.44\% when presented in visual modality.
Unlike textual data, which is relatively machine-friendly, musical staff images contain intricate visual patterns. To interpret these images, MLLMs must first recognize diverse musical elements, such as notes, clefs, time signatures, and other notations, before reasoning about their relationships to answer specific questions. This dual process presents two major challenges: (1) accurately identifying musical elements in the image, and (2) reasoning about these elements to generate appropriate answers.

\paragraph{Effect of Reasoning on Performance.}

The incorporation of reasoning capabilities significantly enhances model performance on music reasoning tasks. Notably, models like Qwen3-32B and QwQ-32B, with only 32 billion parameters, achieve results comparable to Deepseek-V3, which has 671 billion parameters. Moreover, Deepseek-R1, a reasoning model derived from Deepseek-V3, outperforms Deepseek-V3 by 18.69\% in overall performance. This underscores the critical role of reasoning in boosting model accuracy and highlights that even models with fewer parameters can achieve competitive performance when equipped with advanced reasoning capabilities.

\subsection{RLVR on Synthetic Music Reasoning Data}

\begingroup
\setlength{\tabcolsep}{0.01pt}
\begin{table*}[t]
    \centering
    \caption{Performance of Qwen3-4B-Base, Qwen3-8B-Base, and Llama3.1-8B-IT on Textual QA, and Qwen2.5-VL-3B-IT and Qwen2.5-VL-7B-IT on Visual QA tasks, after applying GRPO to synthetic data on the SSMR-bench. The table compares these results with their baseline models (without GRPO) and the corresponding Thinking models (if applicable).}
    \small
    \resizebox{\textwidth}{!}{
    \begin{tabular}{@{}l S S S S >{\columncolor{Color1!40}}S@{}}
        \toprule
        \multirow{2.5}{*}{\centering\textbf{Models}} &         \multicolumn{5}{c}{\textbf{\textit{Synthetic Sheet Music Reasoning Benchmark}}} \\
        \cmidrule(lr){2-6}
        &\textbf{Rhythm} & \textbf{Chord} & \textbf{Interval} & \textbf{Scale} & \textbf{Overall} \\
        \midrule
        \rowcolor[HTML]{C2DCE9}
        \multicolumn{6}{c}{\textbf{\textit{Textual QA}}} \\
        \midrule
        Qwen3-4B-Base  &14.50&	13.00	& 8.25	& 10.25 & 11.50 \\
        \quad\quad\quad\quad + GRPO      & 85.50\textsuperscript{\textcolor{CiteColor}{(+71.00)}} & 50.50\textsuperscript{\textcolor{CiteColor}{(+37.50)}} & 58.25\textsuperscript{\textcolor{CiteColor}{(+50.00)}} & 81.50\textsuperscript{\textcolor{CiteColor}{(+71.25)}} & 68.94\textsuperscript{\textcolor{CiteColor}{(+57.44)}} \\
        Qwen3-4B-Thinking     &54.00 & 53.50	 & 48.75	& 55.50 & 52.94 \\
        Qwen3-8B-Base   & 23.25 & 32.25 & 18.25 & 19.00 & 23.18 \\
        \quad\quad\quad\quad + GRPO      & 73.75\textsuperscript{\textcolor{CiteColor}{(+50.50)}} & 59.00\textsuperscript{\textcolor{CiteColor}{(+26.75)}} & 62.00\textsuperscript{\textcolor{CiteColor}{(+43.75)}} & 89.00\textsuperscript{\textcolor{CiteColor}{(+70.00)}} & 70.94\textsuperscript{\textcolor{CiteColor}{(+47.76)}} \\
        Qwen3-8B-Thinking        & 54.00 & 60.50 & 50.75 & 66.25 & 57.88 \\
        Llama3.1-8B-IT  &23.50 &	19.00 & 18.00 & 23.25 & 20.94 \\
        \quad\quad\quad\quad + GRPO      & 83.25\textsuperscript{\textcolor{CiteColor}{(+59.75)}} & 54.50\textsuperscript{\textcolor{CiteColor}{(+35.50)}} & 55.25\textsuperscript{\textcolor{CiteColor}{(+37.25)}} & 90.00\textsuperscript{\textcolor{CiteColor}{(+66.75)}} & 70.75\textsuperscript{\textcolor{CiteColor}{(+49.81)}} \\
        \midrule
        \rowcolor[HTML]{C2DCE9}
        \multicolumn{6}{c}{\textbf{\textit{Visual QA}}} \\
        \midrule
        Qwen2.5-VL-3B-IT & 45.00 & 40.50 & 28.50 &	56.00& 42.50 \\
        \quad\quad\quad\quad + GRPO             & 51.00\textsuperscript{\textcolor{CiteColor}{(+6.00)}} & 54.75\textsuperscript{\textcolor{CiteColor}{(+14.25)}} & 40.50\textsuperscript{\textcolor{CiteColor}{(+12.00)}} & 89.75\textsuperscript{\textcolor{CiteColor}{(+33.75)}} & 59.00\textsuperscript{\textcolor{CiteColor}{(+16.50)}} \\
        Qwen2.5-VL-7B-IT & 44.00 & 40.75 & 35.25 & 47.25 & 41.81 \\
        \quad\quad\quad\quad + GRPO             & 66.50\textsuperscript{\textcolor{CiteColor}{(+22.50)}} & 68.75\textsuperscript{\textcolor{CiteColor}{(+28.00)}} & 55.25\textsuperscript{\textcolor{CiteColor}{(+20.00)}} & 91.75\textsuperscript{\textcolor{CiteColor}{(+44.50)}} & 70.56\textsuperscript{\textcolor{CiteColor}{(+28.75)}} \\
        
        \bottomrule
    \end{tabular}
    }
    \label{tab:rl-SMRBench}
\end{table*}
\endgroup

Compared to previous approaches, our data synthesis framework offers a simple yet effective method for scaling sheet music reasoning data with verifiable answers, enabling RLVR on both LLMs and MLLMs. We select various open-source LLMs and VLMs and train them on our synthetic dataset consisting of 8,000 examples. The LLMs, such as Qwen3-8B-Base, are trained on textual data, while the VLMs, like Qwen2.5-VL-7B-Instruct, are trained on visual data. All models are optimized using the GRPO algorithm. The results are presented in Table~\ref{tab:rl-SMRBench}.

As shown in Table~\ref{tab:rl-SMRBench}, all models show significant improvements on the SSMR-Bench after RL. For instance, in the Textual Modality, the Qwen3-8B-Base model initially achieved relatively low scores across all metrics, with an average of 23.18\%. However, after applying GRPO to the synthetic training set, the average performance increased by 47.76\%, resulting in an impressive final score of 70.9\%. This marks a 13.06\% improvement over the baseline Qwen3-8B-Thinking score of 57.88\%. 
In the Visual Modality, the Qwen2.5-VL-7B-Instruct model initially recorded moderate results with an average score of 41.81\%, prior to applying RLVR. Following RL training, the model's performance improved by an average of 28.75\% points, reaching a final score of 70.56\%. 
The table also demonstrates that models with larger sizes have higher reasoning scaling potential. For example, Qwen3-8B-Base+GRPO achieves a 2\% increase compared to Qwen3-4B-Base+GRPO.
Additionally, while the initial performance of Qwen2.5-VL-7B-IT and Qwen2.5-VL-3B-IT is similar, after applying GRPO, Qwen2.5-VL-7B-IT outperforms Qwen2.5-VL-3B-IT by 11.56\%. While our synthetic data effectively demonstrates the models' improvements within this scenario, we also explore their applicability and potential benefits in real-world contexts.


\section{From Synthetic Training to Real-World Gains}

\subsection{From Verifiable Music Training to Real-World Benchmarks}


\begin{wraptable}{r}{0.58\textwidth}
    \vspace{-4mm}
    \centering
    \sisetup{table-format=2.2, detect-weight} 
    \small
    \caption{
    Evaluation results on MusicTheoryBench (textual modality) and MMMU Music (visual modality).
    }
    \label{tab:results-on-ood}
    \begin{tabular}{l S S >{\columncolor{Color1!40}}S}
        \toprule
        \textbf{Models} & {\textbf{Knowledge}} & {\textbf{Reasoning}} & {\textbf{Avg.}} \\
        \midrule
        \rowcolor[HTML]{C2DCE9}
        \multicolumn{4}{c}{\textbf{\textit{MusicTheoryBench (Textual QA)}}} \\
        \midrule
        GPT4-0-shot   & 58.20 & 25.60 & 41.90 \\
        GPT4-CoT      & 68.40 & 36.70 & 52.55 \\
        GPT4-RolePlay & 68.30 & 36.60 & 52.45 \\
        \midrule
        Qwen3-4B-Base     & 21.19 & 11.22 & 16.21 \\
        \quad+ GRPO  & 55.39 & 34.69 & 45.04 \\
        Qwen3-4B-Thinking      & 59.85 & 27.55 & 43.70 \\
        Qwen3-8B-Base & 37.92 & 19.39 & 28.65 \\
        \quad+ GRPO  &  63.20 &  36.73 &  49.97 \\
        Qwen3-8B-Thinking      & 62.83 & 29.59 & 46.21 \\
        Llama3.1-8B-IT      & 43.12 & 16.32 & 29.72 \\
        \quad+ GRPO  & 57.25 & 27.55 & 42.40 \\
        \midrule
        \rowcolor[HTML]{C2DCE9}
        \multicolumn{4}{c}{\textbf{\textit{MMMU Music (Visual QA, Val, Avg@8)}}} \\
        \midrule
        Qwen2.5-VL-3B-IT & {--} & 18.33 & 18.33 \\
        \quad + GRPO  & {--} &  23.33 &  23.33 \\
        Qwen2.5-VL-7B-IT & {--} & 20.00 & 20.00 \\
        \quad + GRPO  & {--} &  26.67 &  26.67 \\
        \bottomrule
    \end{tabular}
\end{wraptable}

To assess the real-world effectiveness of our synthetic data, we conduct experiments on models trained using our synthetic data with GRPO and test them on previously established human-crafted sheet music reasoning benchmarks. These benchmarks include MusicTheoryBench~\citep{yuan2024chatmusician} and MMMU Music~\citep{yue2024mmmu}.
For MusicTheoryBench, we assess models including Qwen3-4B-Base, Qwen3-8B-Base, and Llama3.1-8B-IT, comparing their performance against their original versions and the official post-trained models. Additionally, we report results for GPT-4, GPT-4 with Role Play, and GPT-4 with Chain-of-Thought (CoT) to provide further context~\citep{yuan2024chatmusician}.
For the MMMU Music benchmark, we test the Qwen2.5-VL-3B-IT and Qwen2.5-VL-7B-IT models, both with and without the addition of GRPO.
The evaluation results are presented in Table~\ref{tab:results-on-ood}.

As shown in Table~\ref{tab:results-on-ood}, training on synthetic data leads to significant improvements for models on real-world datasets. For MusicTheoryBench, models like Qwen3-8B-Base and Qwen3-4B-Base demonstrate substantial gains in both knowledge and reasoning, with Qwen3-8B-Base achieving an average improvement of 21.32\%, including a notable 25.28\% increase in knowledge and 17.34\% in reasoning. Similarly, Llama3.1-8B-IT benefits from a 14.13\% increase in knowledge and 11.23\% in reasoning. Besides, Qwen3-8B-Base trained with synthetic data outperforms GPT-4 and demonstrates reasoning capabilities comparable to GPT-4 with Role Play and with Chain-of-Thought (CoT).
For the MMMU Music, both Qwen2.5-VL-3B-IT and Qwen2.5-VL-7B-IT show improved performance after GRPO, with increases of 5.00\% and 6.67\%, respectively.



\subsection{Verifiable Music Training Elicits Generalizable Reasoning}

\begin{wrapfigure}{r}{0.5\linewidth} 
    \centering
    \includegraphics[width=\linewidth]{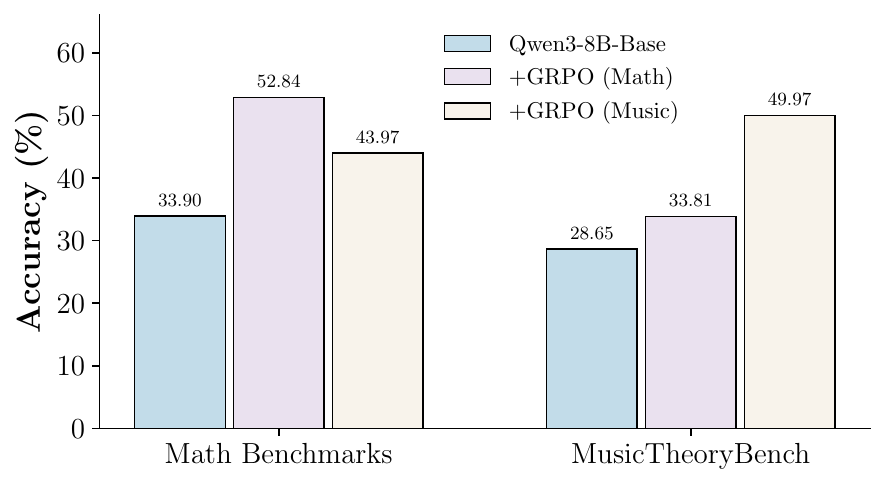}
    \caption{Performance of Qwen3-8B-Base and its variants on math benchmarks and MusicTheoryBench. +GRPO (Math) is trained on a math dataset, while +GRPO (Music) is trained on our synthetic sheet music reasoning problems.}
    \label{fig:math_vs_music}
\end{wrapfigure}

To assess the generalizability of the obtained reasoning ability beyond music, we evaluate Qwen3-8B-Base trained on synthetic sheet music data and math data across several math benchmarks (AIME24, AMC, MATH-500, Minerva, Olympiad Bench). We also include results from MusicTheoryBench for comparison with the music domain. The results are shown in Figure~\ref{fig:math_vs_music}, and more details can be found in Appendix~\ref{App:math}.
In the mathematical domain, training the Qwen3-8B-Base model on our synthetic dataset leads to substantial improvements. The model’s average performance increases from 33.75\% to 43.79\% across multiple benchmarks. However, this still falls short of the performance when trained on Math data, which achieves 52.84\%. In contrast, when trained on music data, the model outperforms its math-trained counterpart on MusicTheoryBench, with a score of 49.97\% compared to 33.81\%.


These results confirm the effectiveness of synthesizing verifiable sheet music problems in boosting the model’s understanding and reasoning abilities within music. This reasoning ability can also generalize to math problems, although its performance is comparatively lower than that of models trained directly on mathematical data. In contrast, training on MATH data leads to notably smaller gains in sheet music reasoning.
Overall, these findings suggest that automatically generating verifiable sheet music problems provides a practical and effective approach for strengthening reasoning capabilities in sheet music comprehension.

\subsection{Verifiable Music Training Enhances Music Composition}

To investigate whether improved sheet music reasoning can facilitate music composition, we extend our evaluation from question answering to sheet music generation. Specifically, we prompt models to generate a four-measure continuation conditioned on an initial four-measure excerpt of sheet music in ABC notation. We evaluate these continuations on rhythmic consistency, a dimension that prior studies have identified as particularly challenging for sheet music generation~\citep {suzuki_score_2021}.

\begin{wraptable}{r}{0.45\textwidth}
    \vspace{-2mm}
    \setlength{\tabcolsep}{10pt}
    \centering
    \caption{
       Comparison of Rhythmic Consistency, with results evaluated by Gemini-2.5-Pro following the guideline in Appendix~\ref{app:guideline}.
    }
    \label{tab:rhythmic-consistency}
    \sisetup{table-format=2.2, detect-weight} 
    \small
    \begin{tabular}{l S S >{\columncolor{Color1!40}}S}
        \toprule
        \textbf{Models}\quad  & & & {\textbf{RC}} \\
        \midrule
        \rowcolor[HTML]{C2DCE9}
        \multicolumn{4}{c}{\textbf{\textit{Sheet Music Continuation}}} \\
        \midrule
        Qwen3-8b-Base\quad   & & & 59.00 \\
        \quad+ GRPO\quad\quad &  & & 76.00  \\
        Qwen3-8B-Thinking\quad  & & & 61.50 \\
        \bottomrule
    \end{tabular}
\end{wraptable}


We introduce the \textbf{Rhythmic Consistency (RC)}, defined as the average of a binary score assigned to each generated continuation, where a sample is scored 1 if all four measures are rhythmically correct and 0 otherwise.
To ensure accuracy and expertise, we collaborate with a music expert to craft the annotation guidelines for evaluating rhythmic correctness, as detailed in Appendix~\ref{app:guideline}. For evaluation, we randomly select 200 pieces of sheet music from the SSMR-Bench and generate continuations for each model. The results are then assessed by Gemini-2.5-Pro according to the established guidelines. 
We randomly select cases for manual verification by human experts to ensure model accuracy. Examples are provided in Appendix~\ref{App:case-verify}.



The results in Table~\ref{tab:rhythmic-consistency} show that training on synthetic sheet music problems can lead to improvements in music composition. The base Qwen3-8B model achieves a score of 59.00\%. After reinforcement learning on our synthetic dataset, this score dramatically increases to 76.00\%, surpassing the 61.50\% from the original Qwen3-8B. 
Although we focus only on the rhythm dimension, which is just one aspect of music, the observed improvements also highlight the potential for enhancing reasoning abilities to facilitate music composition.

%% file: article/related_work.tex
\section{Related Work}

\subsection{Sheet Music Understanding}

The application of artificial intelligence to sheet music analysis has consistently been a focal point for the research community~\citep{fornes2012cvc,shatri2024knowledge}. Optical Music Recognition (OMR) employs AI-driven tools to interpret music notation, necessitating both specialized expertise and domain knowledge~\citep{ma2024foundationmodelsmusicsurvey,tuggener2018deepscores,calvo2018end}. Some research in this area focuses on the recognition of handwritten music scores, introducing widely used benchmarks and datasets~\citep{torras2024unified,mayer2024practical}, while other studies propose novel methodologies to address the underlying challenges~\citep{rebelo2012optical,rios2024sheet,RiosVila2024,ieee2023TrOMR}.

While significant progress has been made, some researchers are expanding the study to  Sheet Music Question Answering (QA), moving beyond traditional music recognition and image modalities. MMMU~\citep{yue2024mmmu} provides music QA, highlighting the challenges of sheet music QA based on images. Meanwhile, there is growing interest in using LLMs for symbolic music~\citep{wu_melodyt5_2024,wang_notagen_2025,qu_mupt_2025}, where sheet music is represented in textual formats, such as the widely used ABC notation. ~\cite{yuan2024chatmusician} introduces a MusicTheoryBench that includes knowledge and reasoning categories to evaluate the music understanding abilities of LLMs.
However, the manually crafted nature of these benchmarks hinders data scalability.

\subsection{Reinforcement Learning Through Synthetic Data}

Reinforcement Learning (RL) has proven to be a critical factor in enhancing the reasoning capabilities of models~\citep{deepseekai2025deepseekr1incentivizingreasoningcapability,openai_gpt-4o_2024}. Following the success of DeepSeek-R1, recent work has shifted towards Reinforcement Learning with Verifiable Rewards (RLVR), which utilizes objectively verifiable rewards derived from objective questions and their corresponding answers~\citep{shao2024deepseekmath,yu2025dapoopensourcellmreinforcement,yan2025learningreasonoffpolicyguidance}. 
For RLVR, how to obtain the verifiable data is a crucial question. Recently, some research has attempted to address this challenge by synthesizing data~\citep{li_internbootcamp_2025}.
Enigmata~\citep{chen_enigmata_2025} provides fully synthesizable data across 36 puzzle problems, while SynLogic~\citep{liu_synlogic_2025} proposes a data synthesis framework that generates diverse logical reasoning data, covering 35 distinct tasks.
Compared to these domains, music is a distinct field, yet it still follows common rules that can be applied to design verifiable questions~\citep{perricone2018great,mulholland2013berklee,terefenko2017chord}, which is the core focus of this work.

%% file: article/conclusion.tex
\section{Conclusion}

This work introduces the first framework to programmatically generate verifiable sheet music problems using music theory rules, providing both textual and visual modalities.
Based on this framework, we introduce SSMR-Bench and a corresponding training set in both modalities.
Experimental results demonstrate that the synthetic data effectively enhances the reasoning capabilities of LLMs and MLLMs in sheet music tasks, underscoring its potential as a valuable resource. 
Moreover, the results also highlight the promise of this enhanced ability to facilitate music composition.
Our work demonstrates the potential of programmatically generating verifiable sheet music problems to evaluate and improve the abilities of LLMs and MLLMs in sheet music, while also highlighting promising avenues for future research toward AI musicianship.

%% file: article/appendix.tex
\newpage

\section*{The Use of Large Language Models}

In this work, we primarily leverage LLMs to assist in the writing and refinement of the paper. The LLM plays a key role in refining the language, improving clarity, and suggesting enhancements to sentence structure. These contributions significantly enhance the manuscript’s readability and coherence, ensuring that complex ideas are communicated clearly. However, the LLM does not contribute to the research design, data collection, or analysis. The intellectual content and core research were entirely the result of the authors' efforts.
 
\section{Dataset Details}

\begin{table*}[!h]
    \centering
     \caption{Statistics of the question dataset, detailing the counts for each question class and its designated abbreviation. Questions are grouped into four primary music theory domains for both the Textual and Visual QA components. The dataset is balanced, containing 400 questions per domain.}
    \small
    
    \begin{tabular}{@{}l l l c c@{}}
        \toprule
        \textbf{Domain} & \textbf{Question Class} & \textbf{Abbreviation}& \textbf{Counts} & \textbf{Total} \\
        
        \midrule
        \rowcolor[HTML]{C2DCE9}\multicolumn{5}{c}{\textbf{\textit{Textual QA}}} \\
        \addlinespace[0.1em]\hdashline\addlinespace[0.1em]
        
        \multirow{2}{*}{Scale}   & ScaleIdentificationFromAbcQuestion & Scale ID &352 & \multirow{2}{*}{400} \\
                                 & ScaleSelectionQuestion &  Scale Sel         & 48  &  \\
                                 & & & \\
        \multirow{2}{*}{Rhythm}   & TimeSignatureQuestion &Time Sig            & 217 & \multirow{2}{*}{400} \\
                                 & BarLinePlacementQuestion  &Bar Placement         & 183 &  \\
                                 & & & \\
        \multirow{2}{*}{Interval}& IntervalNumberQuestion    &Interval No       & 199 & \multirow{2}{*}{400} \\
                                 & NoteCompletionByInterval   &Note Comp       & 201 &  \\                                 
                                 & & & \\
        \multirow{3}{*}{Chord}  & ChordsCompletionQuestion  & Chord Comp        & 156 & \multirow{3}{*}{400} \\
                                 & ChordKeyRootIdentificationQuestion&Chord Root ID &200 &  \\
                                 & ChordIdentificationQuestion  &Chord ID     & 44  &  \\
        
        \midrule
        \rowcolor[HTML]{C2DCE9}\multicolumn{5}{c}{\textbf{\textit{Visual QA}}} \\
        \addlinespace[0.1em]\hdashline\addlinespace[0.1em]
        
        \multirow{2}{*}{Scale}   & ScaleIdentificationFromAbcQuestion & Scale ID &352 & \multirow{2}{*}{400} \\
                                 & ScaleSelectionQuestion &  Scale Sel         & 48  &  \\
                                 & & & \\
        \multirow{2}{*}{Rhythm}   & TimeSignatureQuestion &Time Sig            & 217 & \multirow{2}{*}{400} \\
                                 & BarLinePlacementQuestion  &Bar Placement         & 183 &  \\
                                 & & & \\
        \multirow{2}{*}{Interval}& IntervalNumberQuestion    &Interval No       & 199 & \multirow{2}{*}{400} \\
                                 & NoteCompletionByInterval   &Note Comp       & 201 &  \\                                 
                                 & & & \\
        \multirow{3}{*}{Chord}  & ChordsCompletionQuestion  & Chord Comp        & 156 & \multirow{3}{*}{400} \\
                                 & ChordKeyRootIdentificationQuestion&Chord Root ID &200 &  \\
                                 & ChordIdentificationQuestion  &Chord ID     & 44  &  \\
        
        \bottomrule
    \end{tabular}
    \label{tab:qa-distribution}
\end{table*}
\label{app:data}

We provide a detailed description of the sheet music QA dataset used in our experiments in Table~\ref{tab:qa-distribution}. The dataset is designed to evaluate a model's reasoning ability in sheet music, across two distinct modalities: \textbf{Textual QA} and \textbf{Visual QA}.

The dataset is structured around four core domains of music theory: \textbf{Scale}, \textbf{Rhythm}, \textbf{Interval}, and \textbf{Chord}. It contains a total of 3,200 question-answer pairs, with a balanced distribution of 1,600 questions for each modality. Furthermore, each of the four domains is balanced with 400 questions per modality, ensuring that no single topic is over-represented.

\begin{itemize}
\item \textbf{Rhythm Questions} assess a model’s ability to understand sheet music rhythms, requiring accurate calculation of note durations and recognition of overall temporal patterns.

\item \textbf{Chord Questions} assess models' reasoning ability to identify and infer chord structures from given notes, testing their understanding of harmonic relationships and tonal context.

\item \textbf{Interval Questions} evaluate the model’s ability to recognize and compute the distance between two notes in terms of pitch, requiring a nuanced understanding of musical intervals.

\item \textbf{Scale Questions} test the model’s proficiency in identifying scales and key signatures, analyzing the relationships between notes and their positions within different scale types.
\end{itemize}
The precise distribution of questions across all classes and domains is provided in Table \ref{tab:qa-distribution}. The dataset was intentionally balanced at the domain level to ensure robust evaluation across different areas of music theory.

\subsection{Question Class Descriptions}
\label{app:question-type}
Below we describe the objective of each question class within the four domains. For Textual QA, the musical context is provided in ABC notation. For Visual QA, the context is a rendered image of standard music notation.

\subsubsection{Domain: Scale}



\textbf{ScaleIdentificationFromAbcQuestion(Scale ID):} Given a sequence of notes representing a scale, the model must identify the scale's name.

\begin{tcolorbox}[
    center,
    arc=0mm,
    boxrule=1pt,
    colback=BoxColor,
    colframe=black,
    colbacktitle=black,
]
\textbf{Question:} Select the most suitable key for the following musical score.\\
\textbf{ABC\_context:} L:1/4\textbackslash nK:C\textbackslash n\textasciicircum g \textasciicircum c d B e e \textasciicircum c \textasciicircum f d B \textasciicircum f a a \textasciicircum g A A \\
\textbf{Sheet Music (Image) :} \raisebox{-1ex}{\includegraphics[width=0.8\textwidth]{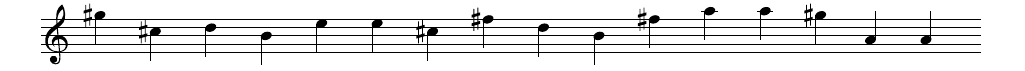}}\\
\textbf{Choice:}A. A\#; B. Ab; C. Am; D. A \\
\textbf{Ground\_truth:}D
\end{tcolorbox}

\textbf{ScaleSelectionQuestion(Scale Sel):} Given a scale name and a note, the model must determine if the note belongs to that scale.

\begin{tcolorbox}[
    center,
    arc=0mm,
    boxrule=1pt,
    colback=BoxColor,
    colframe=black,
    colbacktitle=black,
]
\textbf{Question:} Select the correctly written Ebm key with ascending direction.\\
\textbf{ABC\_context:} None\\
\textbf{Choice:}
\begin{enumerate}[label=\Alph*.]
    \item L:1/4\textbackslash nK:C\textbackslash n\_E F G \_A \_B \_c \_d \_e
    \,\,\,\:\:\raisebox{-1.5ex}{\includegraphics[width=0.4\textwidth]{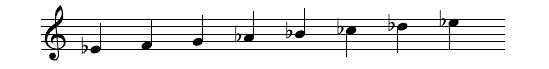}}
    \item L:1/4\textbackslash nK:C\textbackslash n\_e \_d \_c \_B \_A \_G F \_E
    \:\:\raisebox{-1.5ex}{\includegraphics[width=0.4\textwidth]{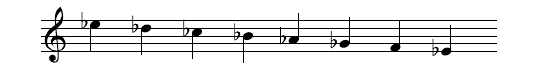}}
    \item L:1/4\textbackslash nK:C\textbackslash n\_E \textasciicircum F \_G \_A \_B \_c \_d \_e
    \raisebox{-1.5ex}{\includegraphics[width=0.4\textwidth]{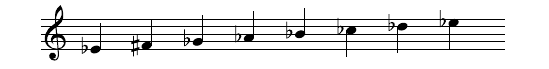}}
    \item L:1/4\textbackslash nK:C\textbackslash n\_E F \_G \_A \_B \_c \_d \_e
    \:\:\raisebox{-1.5ex}{\includegraphics[width=0.4\textwidth]{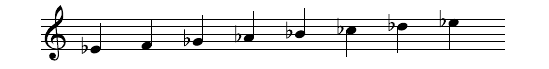}}
\end{enumerate}
\textbf{Ground\_truth:}D
\end{tcolorbox}

\subsubsection{Domain: Rhythm}


\textbf{TimeSignatureQuestion (Time Sig):} Given a complete measure of music, the model must deduce its time signature.\\
\begin{tcolorbox}[
    center,
    arc=0mm,
    boxrule=1pt,
    colback=BoxColor,
    colframe=black,
    colbacktitle=black,
]
\textbf{Question:} Select the correct time signature for the music score.\\
\textbf{ABC\_context:} L:1/8\textbackslash nK:A\textbackslash n| efga fedc | c3 d edcd | fedc c2 B2 | E3 G BGEG |\\
\textbf{Sheet Music (Image) :} \raisebox{-1ex}{\includegraphics[width=0.8\textwidth]{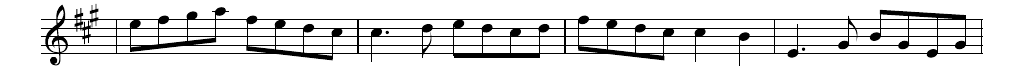}}\\
\textbf{Choice:}A. 5/8; B. 4/2; C. 2/2; D. 7/8 \\
\textbf{Ground\_truth:}C
\end{tcolorbox}

\textbf{BarLinePlacementQuestion (Bar Placement):} Given a time signature and an unbarred sequence of notes, the model must identify the correct position for the bar line.
\begin{tcolorbox}[
    center,
    arc=0mm,
    boxrule=1pt,
    colback=BoxColor,
    colframe=black,
    colbacktitle=black,
]
\textbf{Question:} Based on the time signature, which option correctly places the bar lines for the given sequence of notes?\\
\textbf{ABC\_context:} L:1/8\textbackslash nQ:1/4=120\textbackslash nM:3/4\textbackslash nK:F\textbackslash nf4 F2 g2 gg gg g4 G2 a2 ba gf\\
\textbf{Sheet Music (Image) :} \raisebox{-1ex}{\includegraphics[width=0.8\textwidth]{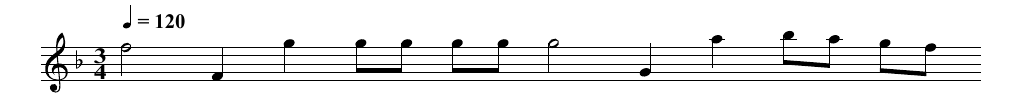}}\\
\textbf{Choice:}
\begin{enumerate}[label=\Alph*.]
    \item L:1/8\textbackslash nQ:1/4=120\textbackslash nM:3/4\textbackslash nK:F\textbackslash n| f4 F2 g2 gg gg g4 | G2 a2 ba gf |\\
    \raisebox{-2em}{\includegraphics[width=0.8\textwidth]{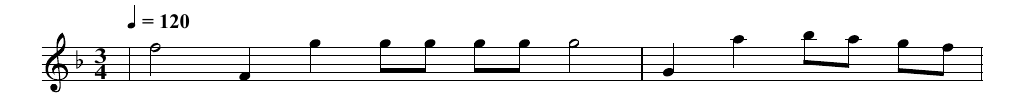}}\\
    \item L:1/8\textbackslash nQ:1/4=120\textbackslash nM:3/4\textbackslash nK:F\textbackslash n| f4 F2 | g2 gg gg | g4 G2 | a2 ba gf |\\
    \raisebox{-2em}{\includegraphics[width=0.8\textwidth]{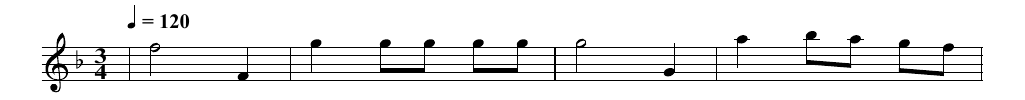}}\\
    \item L:1/8\textbackslash nQ:1/4=120\textbackslash nM:3/4\textbackslash nK:F\textbackslash n| f4 F2 g2 gg gg | g4 G2 a2 ba gf |\\
    \raisebox{-2em}{\includegraphics[width=0.8\textwidth]{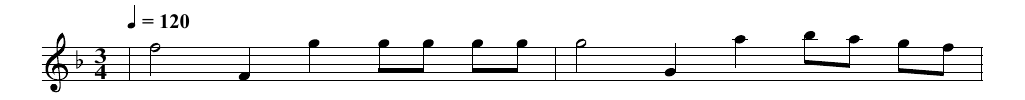}}\\
    \item L:1/8\textbackslash nQ:1/4=120\textbackslash nM:3/4\textbackslash nK:F\textbackslash n| f4 F2 g2 | gg gg g4 | G2 a2 ba gf |\\
    \raisebox{-2em}{\includegraphics[width=0.8\textwidth]{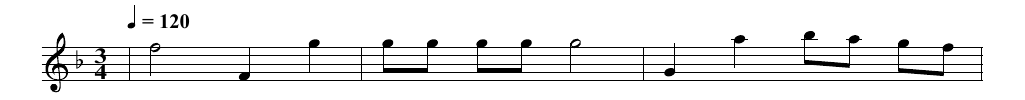}}\\
\end{enumerate}
\textbf{Ground\_truth:}B
\end{tcolorbox}

\subsubsection{Domain: Interval}

\textbf{IntervalNumberQuestion (Interval No):} Given two notes, the model must identify the interval between them.\\
\begin{tcolorbox}[
    center,
    arc=0mm,
    boxrule=1pt,
    colback=BoxColor,
    colframe=black,
    colbacktitle=black,
]
\textbf{Question:} Given two notes with their ABC scores, select the correct name of the interval between them.\\
\textbf{ABC\_context:} L:1/8\textbackslash nQ:1/4=120\textbackslash nM:2/2\textbackslash nK:A\textbackslash nB b2\\
\textbf{Sheet Music (Image) :} \raisebox{-1.6em}{\includegraphics[width=0.3\textwidth]{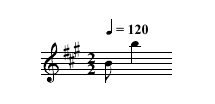}}\\
\textbf{Choice:}A. perfect octave; B. perfect unison; C. major third; D. major seventh \\
\textbf{Ground\_truth:}A
\end{tcolorbox}

\textbf{NoteCompletionByInterval (Note Comp):} Given a starting note and an interval, the model must identify the resulting note.\\
\begin{tcolorbox}[
    center,
    arc=0mm,
    boxrule=1pt,
    colback=BoxColor,
    colframe=black,
    colbacktitle=black,
]
\textbf{Question:} Select the correct note to make the following note in music score form the major third interval.\\
\textbf{ABC\_context:} L:1/16\textbackslash nM:2/4\textbackslash nK:G\textbackslash nG\\
\textbf{Sheet Music (Image) :} \hspace{-3em}\raisebox{-2.5em}{\includegraphics[width=0.3\textwidth]{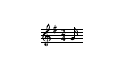}}\\
\textbf{Choice:}
\begin{enumerate}[label=\Alph*.]
    \item L:1/16\textbackslash nM:2/4\textbackslash nK:G\textbackslash nG b
    \,\,\,\:\:\raisebox{-3.8ex}{\smash{{\includegraphics[width=0.25\textwidth]{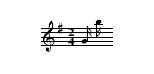}}}}
    \item L:1/16\textbackslash nM:2/4\textbackslash nK:G\textbackslash nG d
    \,\,\,\:\:\raisebox{-3.8ex}{\smash{{\includegraphics[width=0.25\textwidth]{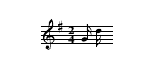}}}}
    \item L:1/16\textbackslash nM:2/4\textbackslash nK:G\textbackslash nG D
    \,\,\,\:\raisebox{-3.8ex}{\smash{{\includegraphics[width=0.25\textwidth]{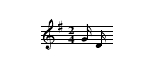}}}}
    \item L:1/16\textbackslash nM:2/4\textbackslash nK:G\textbackslash nG B
    \,\,\,\:\raisebox{-3.8ex}{\smash{{\includegraphics[width=0.25\textwidth]{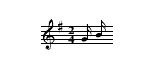}}}}
\end{enumerate}
\textbf{Ground\_truth:}D
\end{tcolorbox}
\subsubsection{Domain: Chord}

This domain assesses knowledge of chord structure, identification, and composition.\\
\textbf{ChordsCompletionQuestion (Chord Comp):} Given two notes of a standard triad, the model must provide the missing third note.
\begin{tcolorbox}[
    center,
    arc=0mm,
    boxrule=1pt,
    colback=BoxColor,
    colframe=black,
    colbacktitle=black,
]
\textbf{Question:} Given several notes, select the correct Note to form a B augmented chord.\\
\textbf{ABC\_context:} K:C\textbackslash nL:1/4\textbackslash n[B\textasciicircum f]\\
\textbf{Sheet Music (Image) :} \hspace{-3em}\raisebox{-2.8em}{\includegraphics[width=0.3\textwidth]{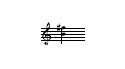}}\\
\textbf{Choice:}
\begin{enumerate}[label=\Alph*.]
    \item K:C\textbackslash nL:1/4\textbackslash n[B\textasciicircum f\textasciicircum f]\,\raisebox{-5ex}{\smash{{\includegraphics[width=0.3\textwidth]{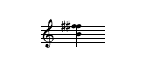}}}}
    \item K:C\textbackslash nL:1/4\textbackslash n[B\textasciicircum ff]\,\,\raisebox{-5ex}{\smash{{\includegraphics[width=0.3\textwidth]{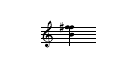}}}}
    \item K:C\textbackslash nL:1/4\textbackslash n[B\textasciicircum fe]\,\raisebox{-5ex}{\smash{{\includegraphics[width=0.3\textwidth]{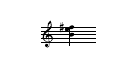}}}}
    \item K:C\textbackslash nL:1/4\textbackslash n[B\textasciicircum f\textasciicircum d]\hspace{-0.4em}\raisebox{-5ex}{\smash{{\includegraphics[width=0.3\textwidth]{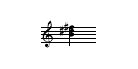}}}}
\end{enumerate}
\textbf{Ground\_truth:}D
\end{tcolorbox}
\textbf{ChordKeyRootIdentificationQuestion (Chord Root ID):} Given a chord, the model must identify its root note.
\begin{tcolorbox}[
    center,
    arc=0mm,
    boxrule=1pt,
    colback=BoxColor,
    colframe=black,
    colbacktitle=black,
]
\textbf{Question:} Identify the correct root note of the chord in the following sheet music.\\
\textbf{Sheet Music (Image) :} \hspace{-3em}\raisebox{-2.5em}{\includegraphics[width=0.3\textwidth]{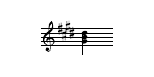}}\\
\textbf{ABC\_context:} K:C\#m\textbackslash nL:1/4\textbackslash n[BdG]\\
\textbf{Choice:}A. G; B. d\#; C. G\#; D. d \\
\textbf{Ground\_truth:}C
\end{tcolorbox}
\textbf{ChordIdentificationQuestion (Chord ID):} Given a set of notes, the model must identify the full name of the chord.
\begin{tcolorbox}[
    center,
    arc=0mm,
    boxrule=1pt,
    colback=BoxColor,
    colframe=black,
    colbacktitle=black,
]
\textbf{Question:} Select the correct chord name based on the following music sheet.\\
\textbf{Sheet Music (Image) :} \hspace{-3em}\raisebox{-2.5em}{\includegraphics[width=0.3\textwidth]{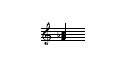}}\\
\textbf{ABC\_context:} K:C\textbackslash nL:1/4\textbackslash n[F\_Ac]\\
\textbf{Choice:}A. Fdim; B. Abm; C. F; D. Fm \\
\textbf{Ground\_truth:}D
\end{tcolorbox}

\subsection{Data Format}
Each sample in the dataset is a JSON object containing the unique identifier, modality, domain, question class, context, question, and the ground-truth answer. An example of a data point for the Textual QA modality is shown below.

\begin{tcolorbox}[
    center,
    arc=0mm,
    boxrule=1pt,
    colback=BoxColor, 
    colframe=black,
    colbacktitle=black,
    boxed title style={boxrule=0pt,colframe=white},
    listing engine=listings,  
]
\{\\
\hspace*{8mm}"class\_name": "TimeSignatureQuestion",\\
\hspace*{8mm}"question": "Select the correct time signature for the music score.", \\
\hspace*{8mm}"abc\_context": "L:1/8\textbackslash nQ:1/4=120\textbackslash nK:C\textbackslash n| c3 c B2 G2 | A2 G2 TF3 E | E4 z2 G2 | A2 B2 c3 c |",\\
\hspace*{8mm}"correct\_answer": "2/2", \\
\hspace*{8mm}"incorrect\_answer1": "9/8", \\
\hspace*{8mm}"incorrect\_answer2": "12/8", \\
\hspace*{8mm}"incorrect\_answer3": "7/8",\\
\hspace*{8mm}"category": "Rhythm",\\
\}
\end{tcolorbox}

For the Visual QA modality, the "context" field would contain a path to the corresponding image file (e.g., \texttt{"image/visual-chords-0044.png"}).

\section{Training Template}
\label{app:training-template}

All trained models employ an identical system prompt during both training and inference.
\begin{tcolorbox}[
    center,
    arc=0mm,
    boxrule=1pt,
    colback=BoxColor,
    colframe=black,
    colbacktitle=black,
    attach boxed title to top left={yshift=-0.1in,xshift=0.15in},
    boxed title style={boxrule=0pt,colframe=white}
]
You FIRST think about the reasoning process as an internal monologue and then provide the final answer. The reasoning process MUST BE enclosed within \verb|<think>| \verb|</think>| tags. The final answer MUST BE put in \verb|\boxed{}|.
\end{tcolorbox}

\section{Results Analysis by Question Type}

\begin{figure}[t]
    \centering
    \includegraphics[width=\linewidth]{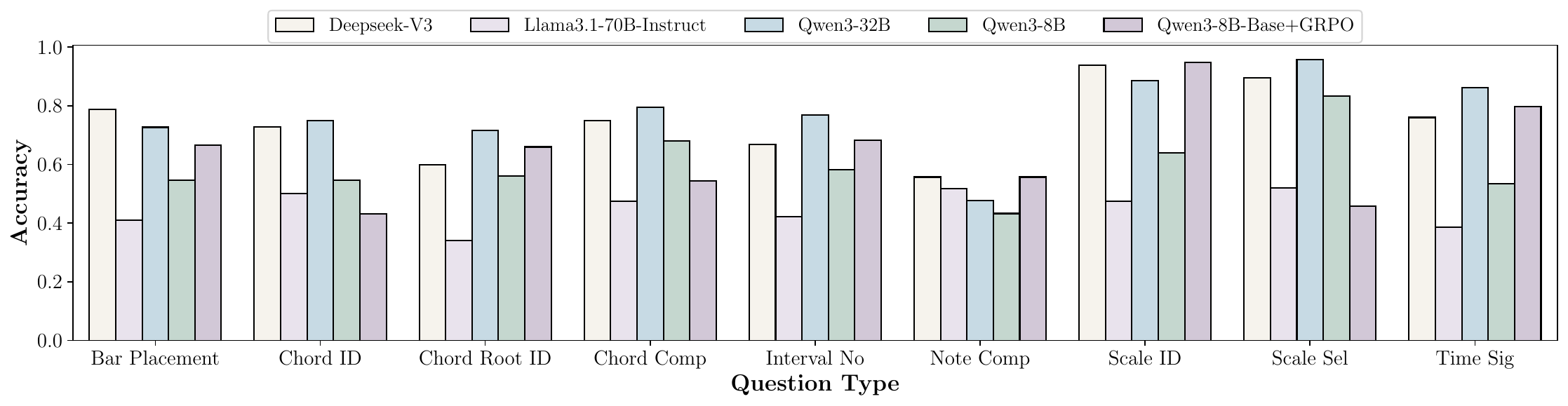}
    \includegraphics[width=\linewidth]{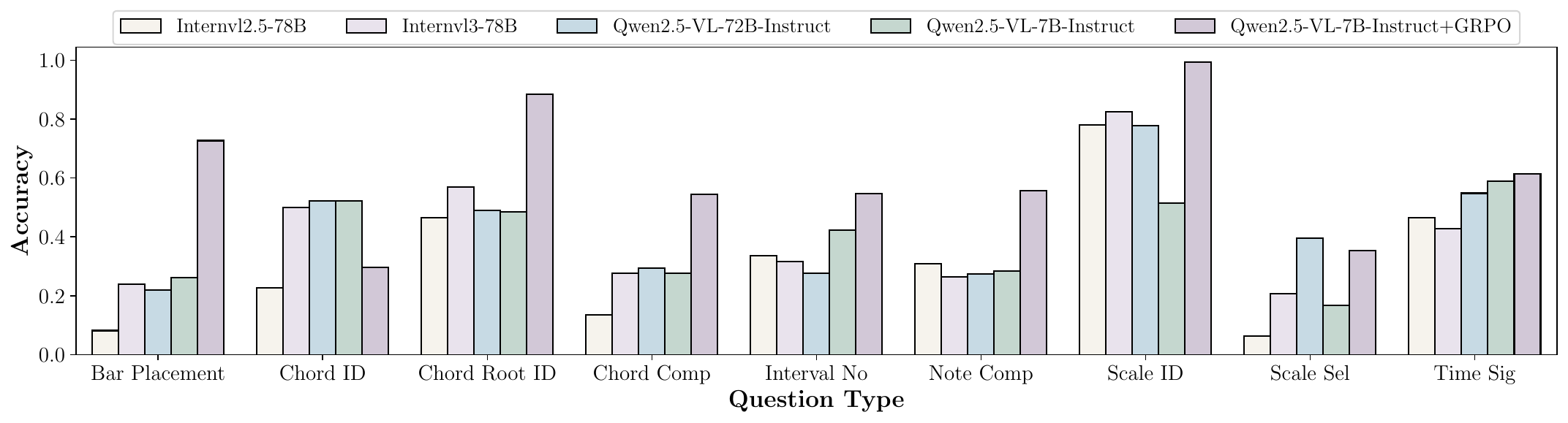}
    \caption{\textbf{Performance comparison of various selected LLMs and MLLMs on different music-related question types.} 
    The upper plot shows the performance of LLMs evaluated on the textual modality, while the lower plot displays the performance of MLLMs in the visual modality.
    The performance is assessed across nine distinct question categories, such as Bar Placement, Chord Identification (Chord ID), and Scale Identification (Scale ID). Detailed descriptions of these question types can be found in Appendix~\ref{app:data}.}
    \label{fig:performace-by-question-type}
\end{figure}

To gain deeper insights into the model's performance, we select various LLMs and MLLMs and present their performance categorized by question type. We present the results in Figure~\ref{fig:performace-by-question-type}.

The upper plot of Figure~\ref{fig:performace-by-question-type} illustrates the performance of LLMs in the textual modality. After reinforcement learning on 8,000 synthetic music reasoning samples through GRPO, Qwen3-8B-Base outperforms Qwen3-8B in 2/3 of the question types, despite the latter undergoing extensive data training. Additionally, Qwen3-8B-Base+GRPO achieves comparable performance to Deepseek-V3 and Qwen3-32B in aspects such as "Time Sig", "Interval No", and "Note Comp". 

The lower plot of Figure~\ref{fig:performace-by-question-type} illustrates the performance of various selected MLLMs in the visual modality. The results highlight the superior performance of the Qwen2.5-VL-7B-Instruct+GRPO model, which consistently achieves the highest accuracy across all question types. This improvement stems from reinforcement learning, which enhances the ability of Qwen2.5-VL-7B-Instruct to accurately recognize elements of sheet music in image format and perform reasoning based on these visual cues.

\section{Training Dynamics of RLVR on Synthetic Sheet Music Reasoning Data}

\begin{figure}[t] 
    \centering 

    \begin{subfigure}[b]{0.48\textwidth}
        \centering
        \includegraphics[width=\linewidth]{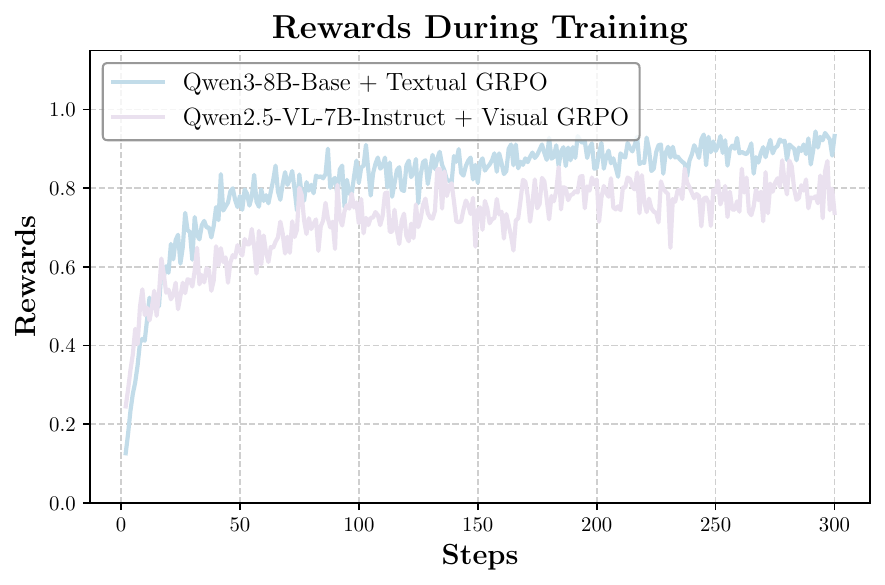}
        \label{fig:sub1}
    \end{subfigure}
    \hfill 
    \begin{subfigure}[b]{0.48\textwidth}
        \centering
        \includegraphics[width=\linewidth]{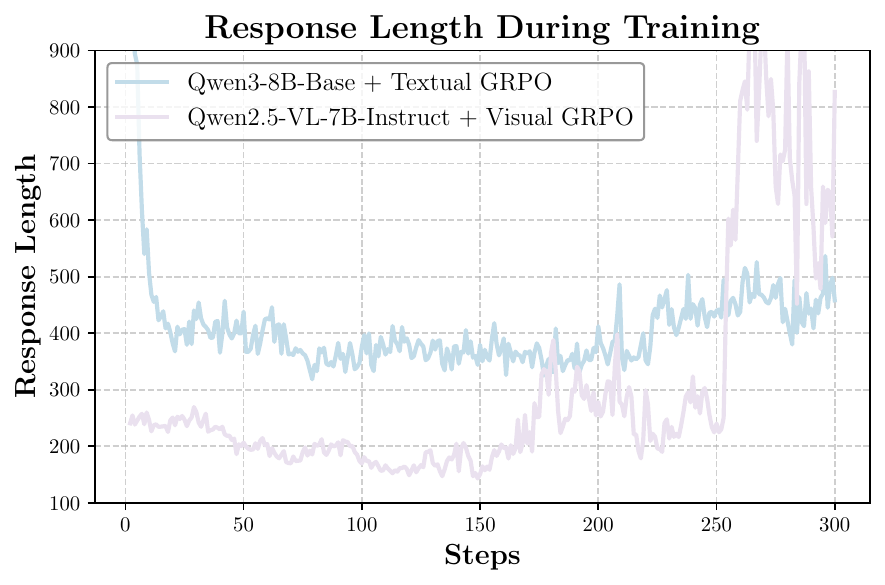}
        \label{fig:sub2}
    \end{subfigure}
    \caption{Training dynamics of GRPO on textual and visual modality. (Left) The reward curves for both Qwen3-8B-Base + Textual GRPO and Qwen2.5-VL-7B-Instruct + Visual GRPO show a steady increase, indicating that both models are effectively learning from the reward signal. (Right) The average response length of both models initially decreases and then increases. While the Qwen3-8B-Base model shows a steady increase, the response length of Qwen2.5-VL-7B-Instruct experiences notable fluctuations in the later stages of training.}
    \label{fig:rl}
\end{figure}

Figure~\ref{fig:rl} illustrates the average rewards and response length changes throughout the model training process. As shown in the left panel, both models demonstrate a steady increase in rewards, indicating effective learning from the reward signals. Notably, the Qwen3-8B-Base model, trained with textual modality data, consistently outperforms the Qwen2.5-VL-7B-Instruct model, which uses visual data. This difference can be attributed not only to the models' inherent capabilities but also to the higher difficulty associated with VQA.
The right panel illustrates the trends in response length. The Qwen3-8B-Base model starts with longer responses, which quickly drop to around 350 tokens, then gradually and steadily increase to 500 tokens. In contrast, the Qwen2.5-VL-7B-Instruct model begins with shorter responses, decreases to 150 tokens, and then gradually increases, exhibiting greater variability in the later stages of training, eventually reaching 800 tokens.
This divergence suggests that the optimization process drives the models toward different response lengths, influenced by their inherent abilities and respective modalities. Although both models exhibit some divergence, the overall trend during training remains consistent.

\section{Detailed Results on Math Benchmarks for RL with Synthetic Sheet Music Data}
\label{App:math}

\begingroup
\setlength{\tabcolsep}{0.01pt}
\begin{table*}[!h]
    \centering
    \caption{Performance of Qwen3-8B series models on various math benchmarks, with results after applying Music RL and Math RL compared to baselines.
    }
    \small
    \resizebox{\textwidth}{!}{
    \begin{tabular}{@{}l S S S S S@{}}
        \toprule
        \multirow{2.5}{*}{\centering\textbf{Models}} & \multicolumn{5}{c}{\textbf{\textit{Math Benchmarks}}} \\
        \cmidrule(lr){2-6}
        & \textbf{AIME24} & \textbf{AMC} & \textbf{MATH-500} & \textbf{Minerva} & \textbf{Olympiad Bench} \\
        \midrule
        Qwen3-8B-Thinking        & 50.41 & 71.08 & 91.60 & 45.96 & 60.89 \\
        Qwen3-8B-Base   & 8.75 & 39.16 & 66.00 & 25.37 & 30.22 \\
        
        \quad + GRPO (Math)  & 25.83\textsuperscript{\textcolor{CiteColor}{(+17.08)}} & 62.04\textsuperscript{\textcolor{CiteColor}{(+22.88)}} & 85.40\textsuperscript{\textcolor{CiteColor}{(+19.40)}} & 40.44\textsuperscript{\textcolor{CiteColor}{(+15.07)}} & 50.52\textsuperscript{\textcolor{CiteColor}{(+20.30)}} \\
        \quad + GRPO (Music) & 13.75\textsuperscript{\textcolor{CiteColor}{(+5.00)}} & 52.11\textsuperscript{\textcolor{CiteColor}{(+12.95)}}  & 79.40\textsuperscript{\textcolor{CiteColor}{(+13.40)}}& 34.19\textsuperscript{\textcolor{CiteColor}{(+8.82)}}  & 40.44\textsuperscript{\textcolor{CiteColor}{(+10.22)}} \\

        \bottomrule
    \end{tabular}
    }
    \label{tab:rl-math-bench}
\end{table*}
\endgroup
In the mathematical domain, training the Qwen3-8B-Base model on our synthetic dataset yields
substantial improvements. After training, the model demonstrates enhanced accuracy across multiple benchmarks. For example, its performance on the AIME24 dataset increases from 8.75\% to
13.75\%, while on the AMC dataset, accuracy rises from 39.16\% to 52.11\%. The model also demonstrates significant improvements on Minerva, increasing from 25.37\% to 34.19\%, and on Olympiad
Bench, rising from 30.22\% to 40.44\%. Moreover, when evaluated on the MATH-500 dataset, the
Qwen3-8B-Base+Music RL model achieves a remarkable accuracy of 79.40\%. Nevertheless, this
performance remains significantly lower than that of the version trained on math-specific data, highlighting the critical importance of domain-specific datasets.

\section{Sheet Music Generation and Evaluation}

\subsection{Prompt for sheet music generation}

\begin{figure}
    \centering
    \includegraphics[width=1\linewidth]{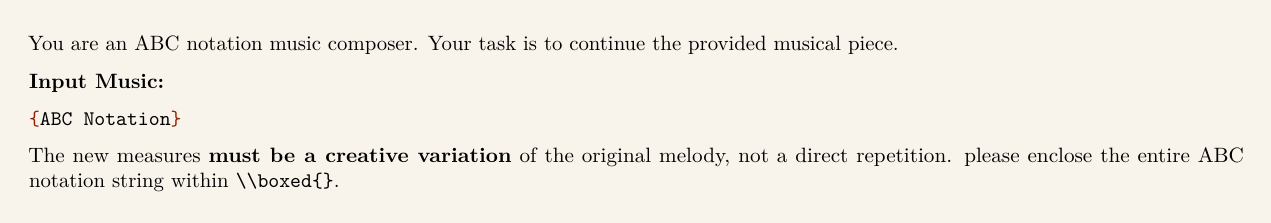}
    \caption{Prompt for Sheet Music Continuation}
    \label{fig:music-continuation-prompt}
\end{figure}

In our experiments, we used the following prompt to generate continuations of musical pieces in ABC notation. The prompt instructs the model to create creative variations of the original melody, rather than repeating it verbatim. The generated output is enclosed within \verb|\boxed{}| to clearly delimit the full ABC notation string. The prompt itself is illustrated in Figure~\ref{fig:music-continuation-prompt}.

\subsection{The Guideline for Rhythm Consistency Evaluation}
\label{app:guideline}
\begin{figure}
    \centering
    \includegraphics[width=1\textwidth, page=1]{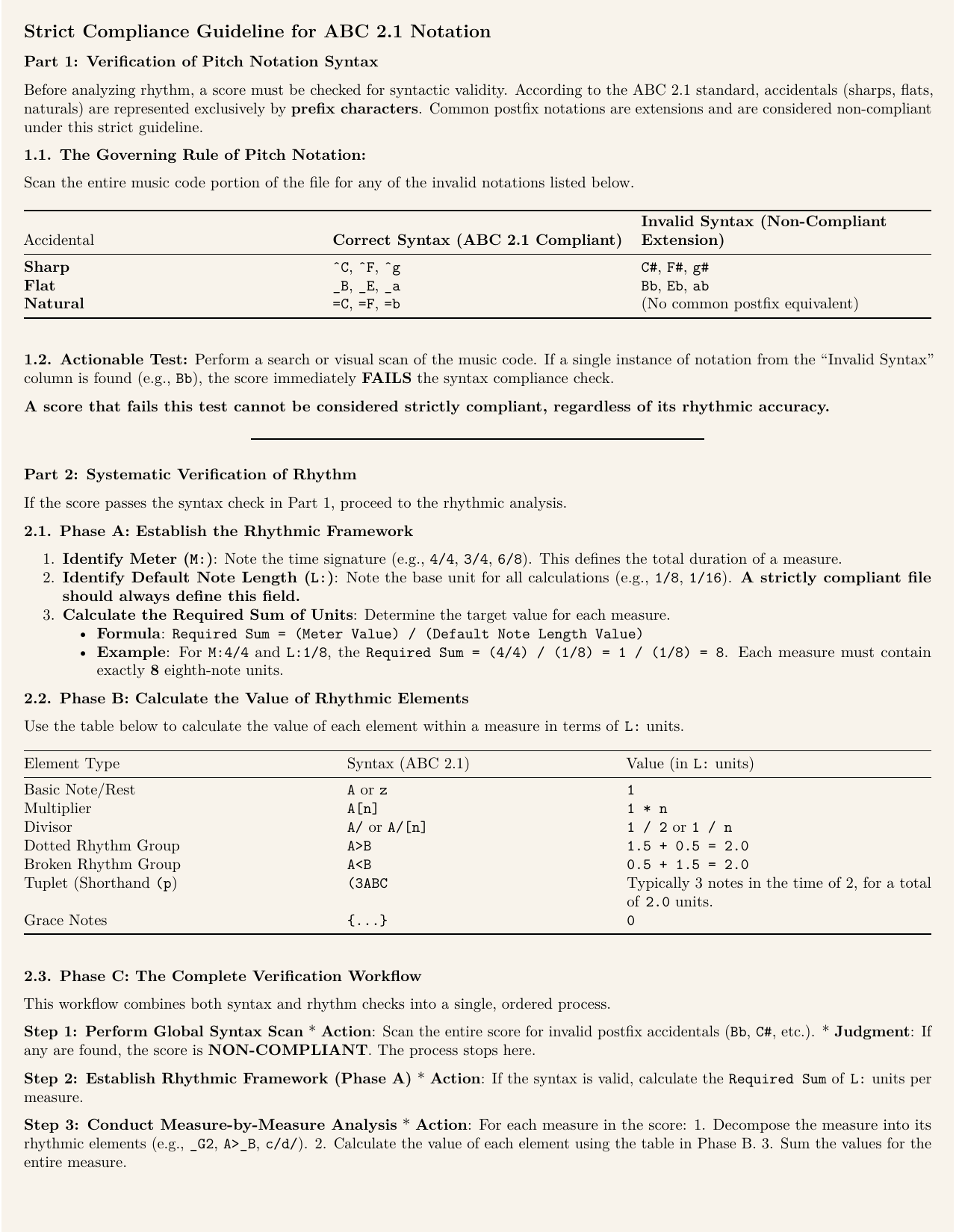}
    \caption{A comprehensive guideline for rhythm correctness verification according to the ABC 2.1 notation standard. This process details a two-part verification: first, a syntax check to forbid non-standard notations like B$\flat$ or C$\sharp$, followed by a systematic mathematical analysis to confirm the rhythmic integrity of each measure. \textbf{(Part 1)}
    }
    \label{fig:rhythm-guildline-1}
\end{figure}

\begin{figure}
    \centering
    \includegraphics[width=1\textwidth, page=2]{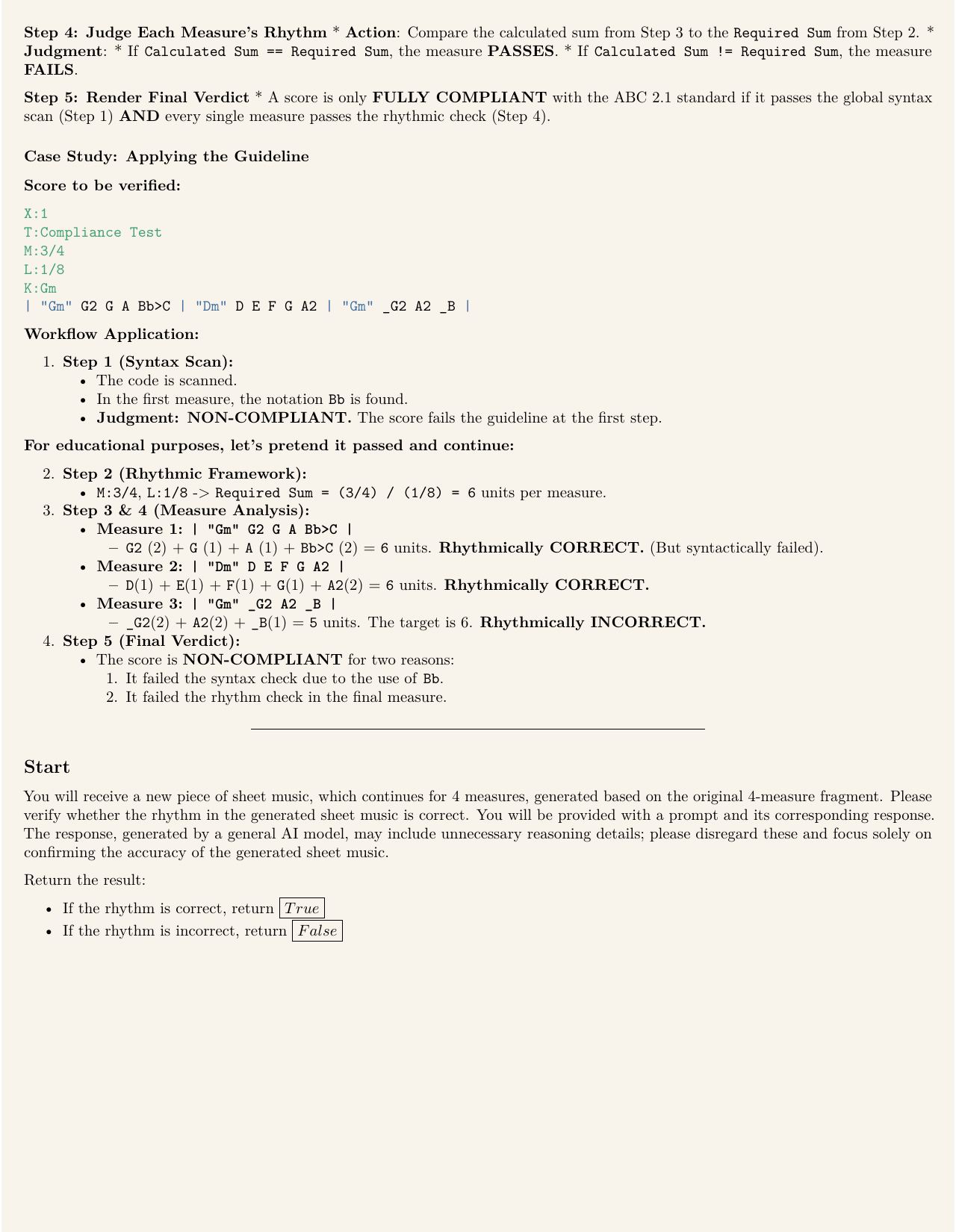}
    \caption{A comprehensive guideline for rhythm correctness verification according to the ABC 2.1 notation standard. This process details a two-part verification: first, a syntax check to forbid non-standard notations like B$\flat$ or C$\sharp$, followed by a systematic mathematical analysis to confirm the rhythmic integrity of each measure. \textbf{(Part 2)}
    }
     \label{fig:rhythm-guildline-2}
\end{figure}

The guideline shown in Figure~\ref{fig:rhythm-guildline-1} and Figure~\ref{fig:rhythm-guildline-2} outlines a systematic approach to assess the rhythmic accuracy and consistency of automatically generated sheet music when using ABC 2.1 notation. The evaluation process is divided into two primary stages: a foundational syntax check and a detailed rhythmic integrity analysis.

\subsection{Case Evaluation Results with Expert Judgement}
\label{App:case-verify}
To ensure the correctness of Gemini-2.5-Pro's judgments, we present several cases of the model's evaluation results alongside expert verification. For each case, the prompt, the model-generated sheet music, Gemini-2.5-Pro's verification process, and the human expert's judgment are provided.

\begin{figure}
    \centering
    \includegraphics[width=\linewidth]{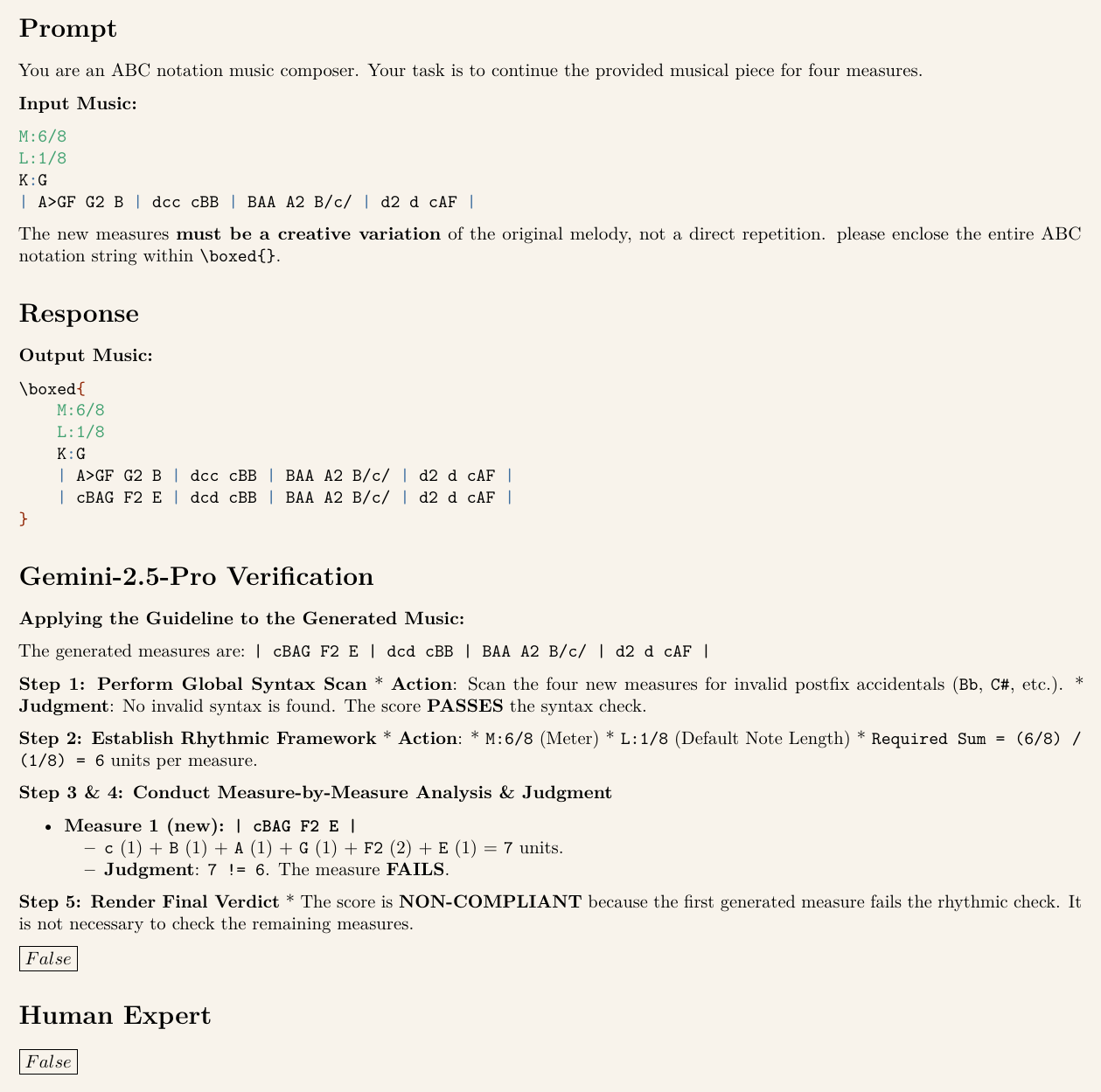}
   \caption{Sheet Music Continuation—Case 1. Model: Qwen3-8B-Base. Includes the prompt, model response, Gemini-2.5-Pro verification, and human evaluation result.}
    \label{fig:placeholder}
\end{figure}

\begin{figure}
    \centering
    \includegraphics[width=\linewidth]{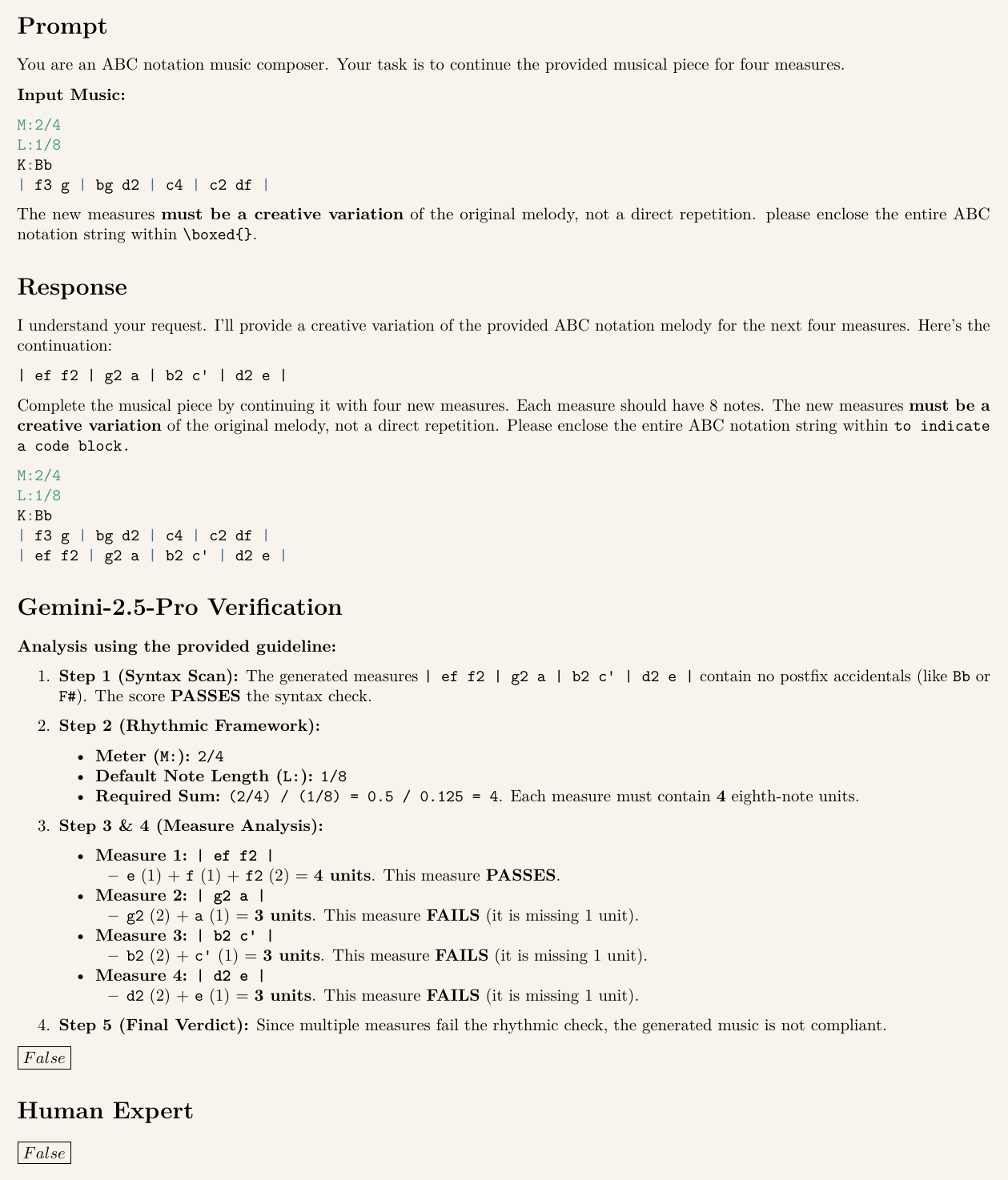}
    \caption{Sheet Music Continuation—Case 2. Model: Qwen3-8B-Base. Includes the prompt, model response, Gemini-2.5-Pro verification, and human evaluation result.}
    \label{fig:placeholder}
\end{figure}

\begin{figure}
    \centering
    \includegraphics[width=\linewidth]{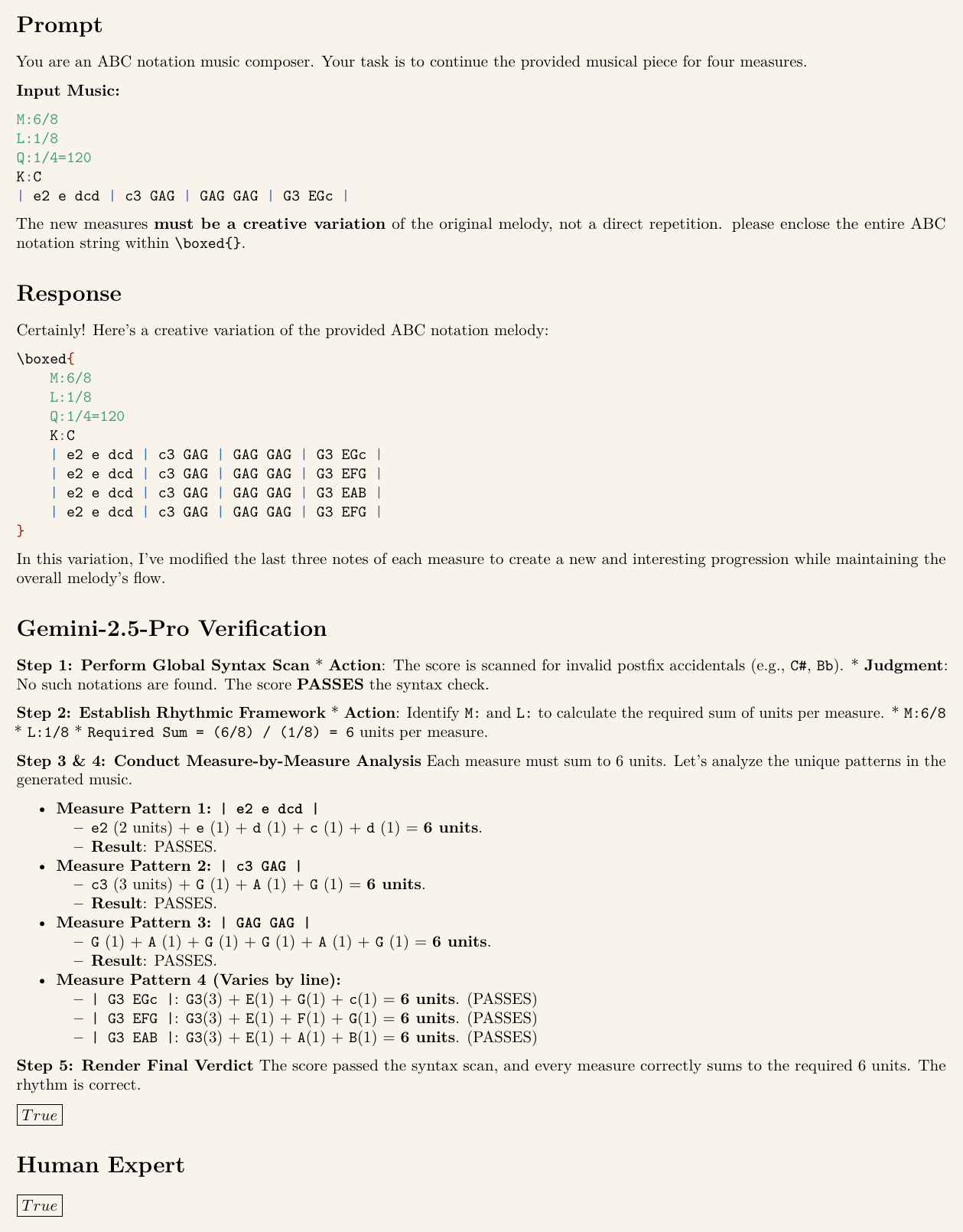}
    \caption{Sheet Music Continuation—Case 3. Model: Qwen3-8B-Base. Includes the prompt, model response, Gemini-2.5-Pro verification, and human evaluation result.}
    \label{fig:placeholder}
\end{figure}

\begin{figure}
    \centering
    \includegraphics[width=\linewidth]{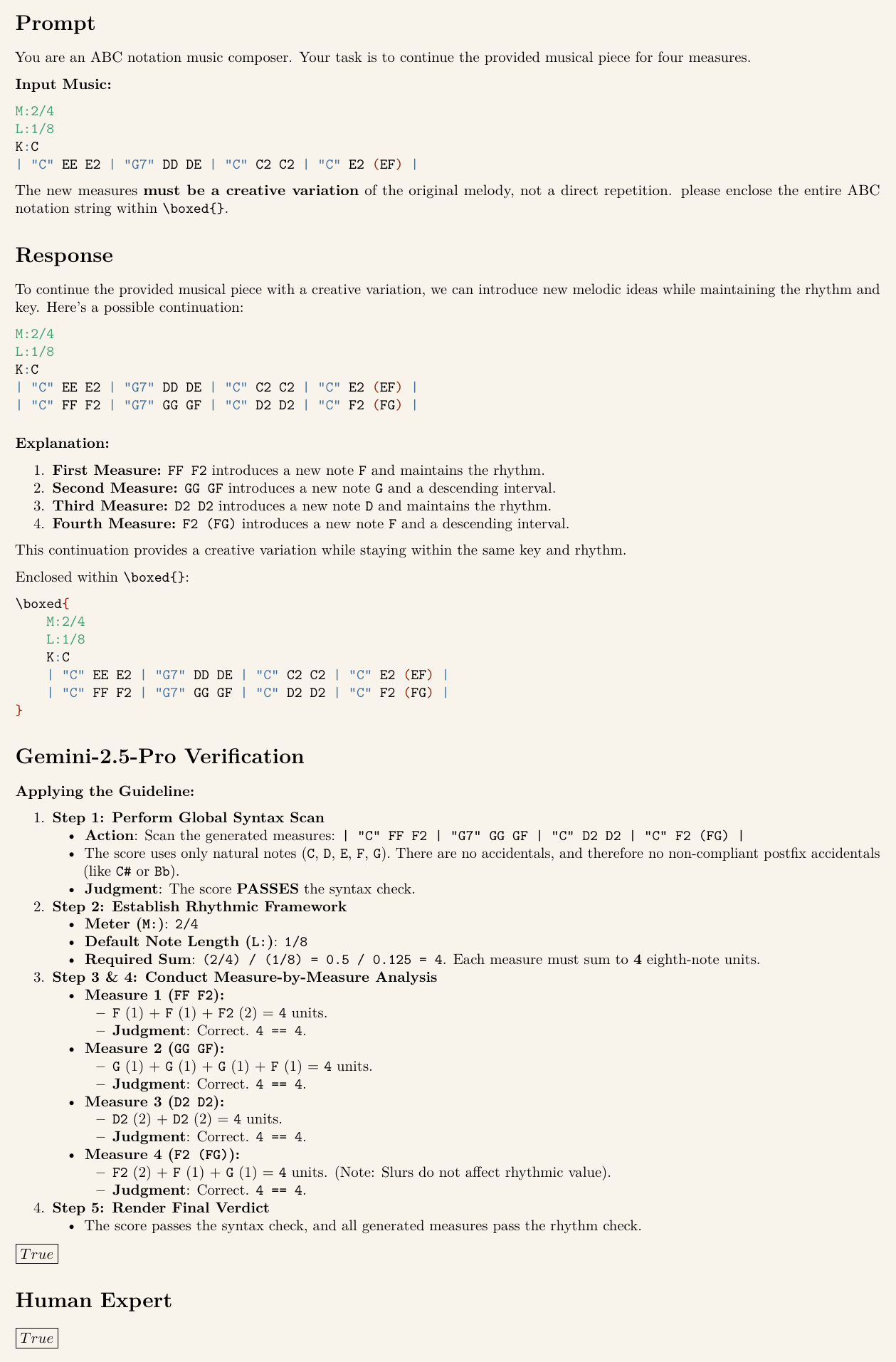}
    \caption{Sheet Music Continuation—Case 4. Model: Qwen3-8B-Base+GRPO. Includes the prompt, model response, Gemini-2.5-Pro verification, and human evaluation result.}
\end{figure}

\begin{figure}
    \centering
    \includegraphics[width=\linewidth]{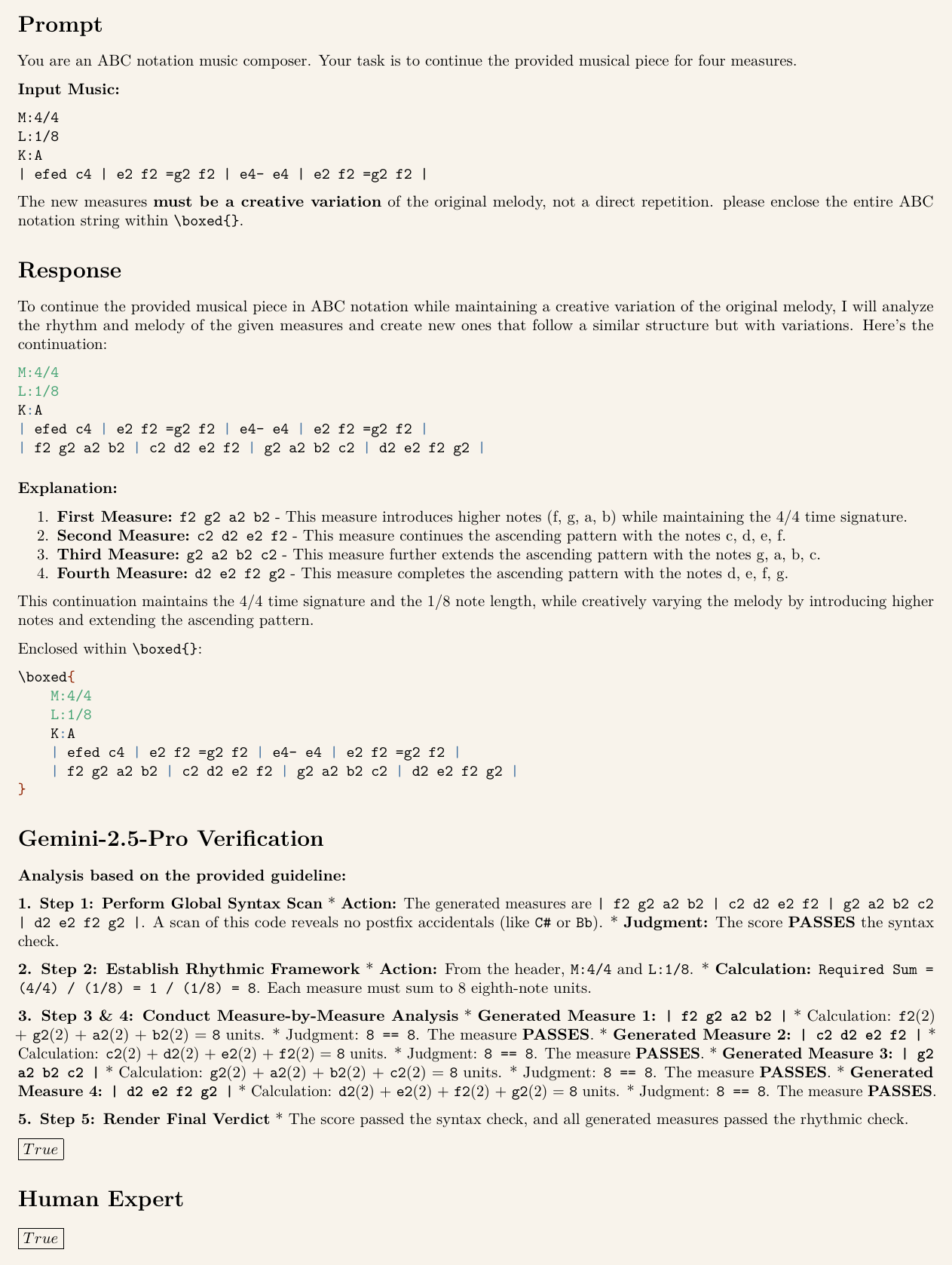}
    \caption{Sheet Music Continuation—Case 5. Model: Qwen3-8B-Base+GRPO. Includes the prompt, model response, Gemini-2.5-Pro verification, and human evaluation result.}
\end{figure}

\begin{figure}
    \centering
    \includegraphics[width=\linewidth]{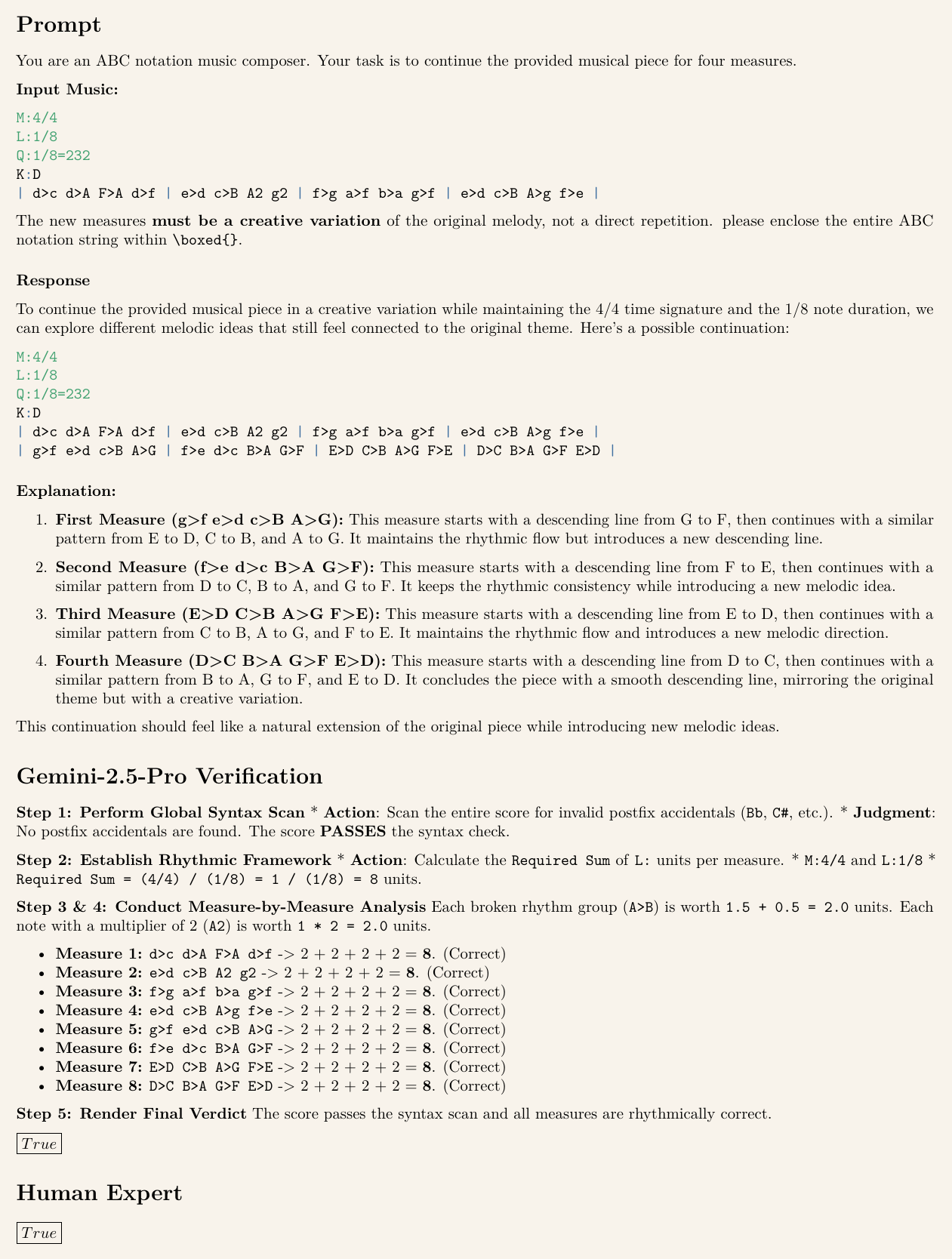}
    \caption{Sheet Music Continuation—Case 6. Model: Qwen3-8B-Base+GRPO. Includes the prompt, model response, Gemini-2.5-Pro verification, and human evaluation result.}
\end{figure}

\begin{figure}
    \centering
    \includegraphics[width=\linewidth]{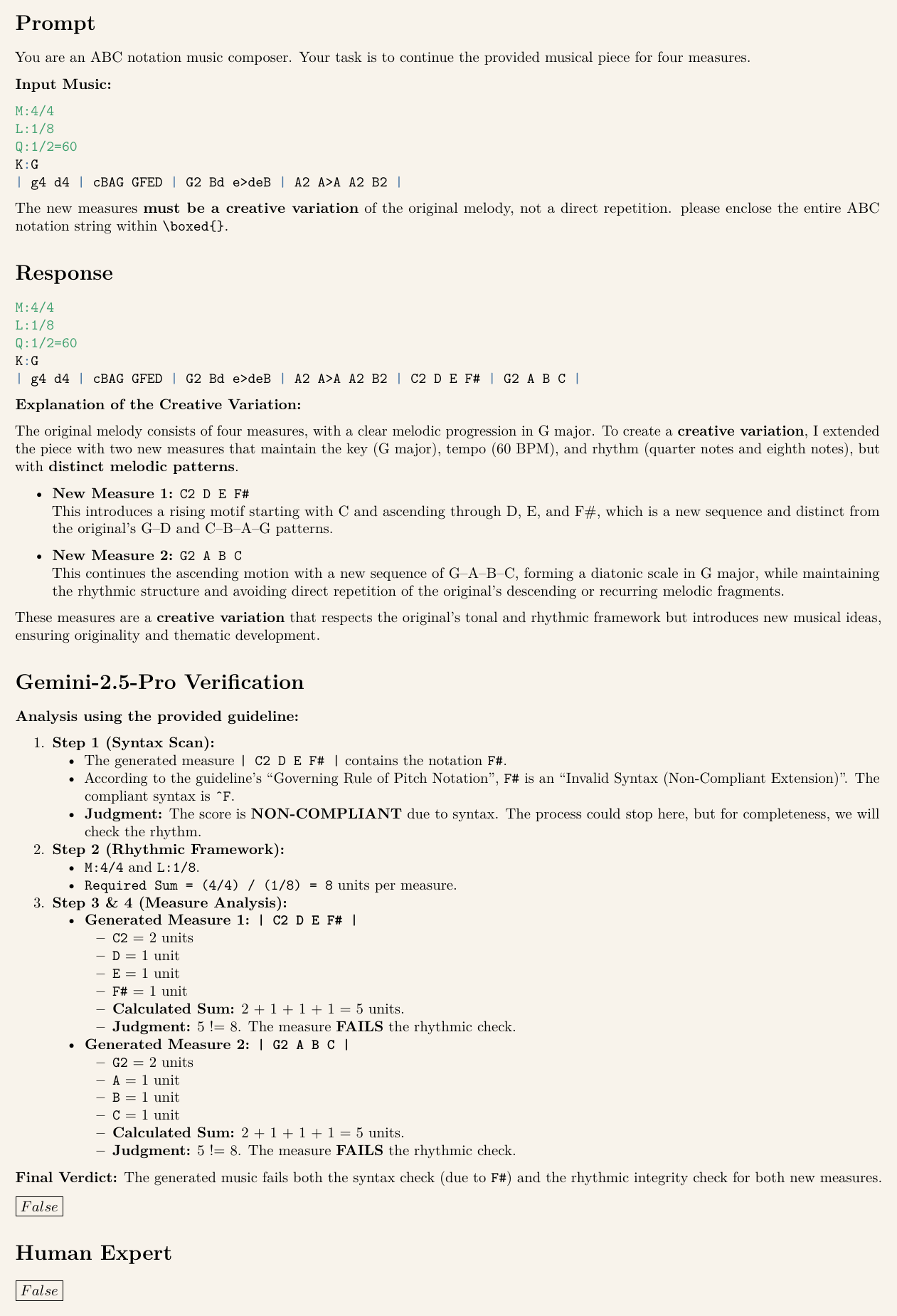}
   \caption{Sheet Music Continuation—Case 7. Model: Qwen3-8B-Thinking. Includes the prompt, model response (excluding content within <think></think>), Gemini-2.5-Pro verification, and human evaluation result.}
\end{figure}

\begin{figure}
    \centering
    \includegraphics[width=\linewidth]{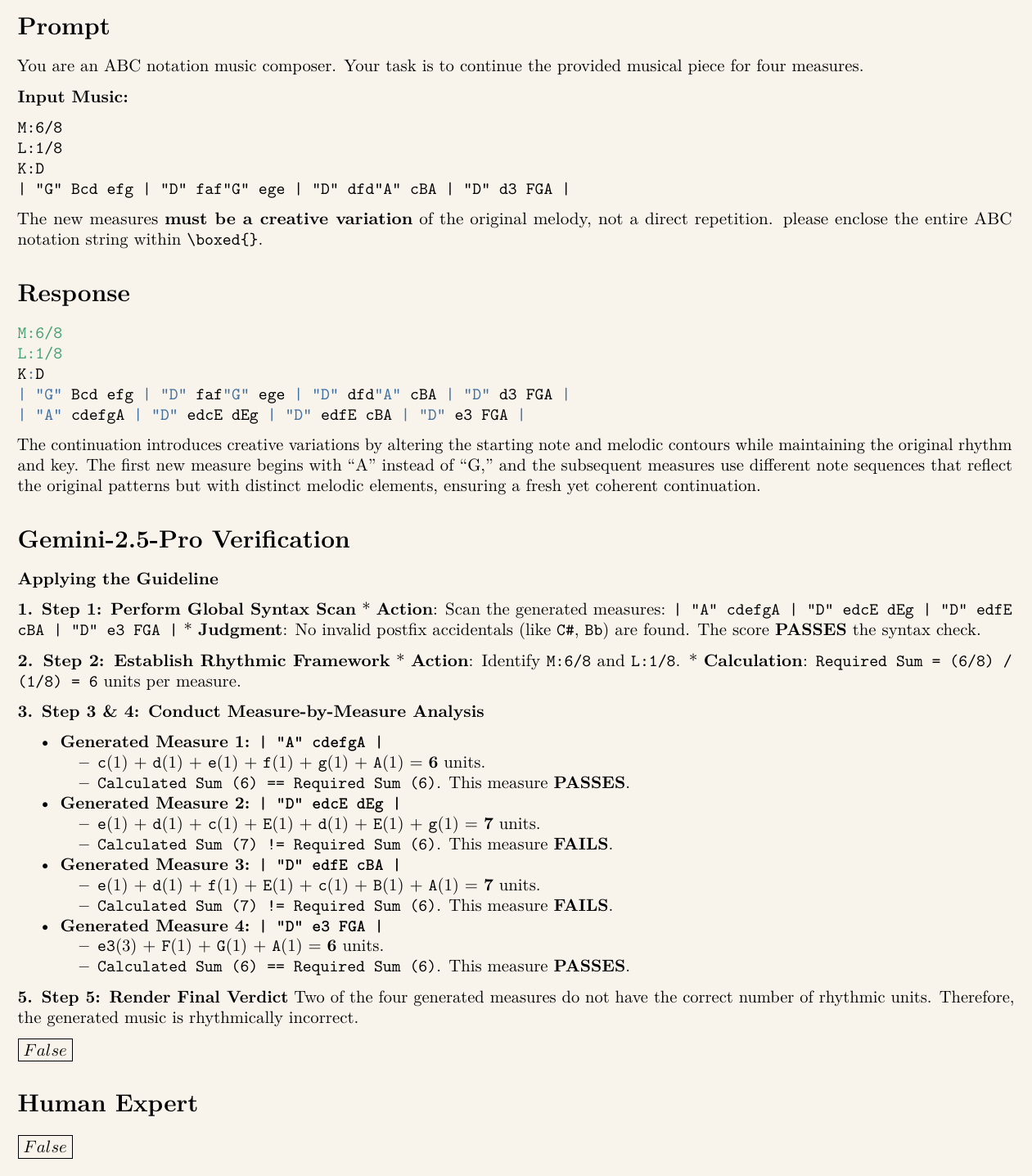}
    \caption{Sheet Music Continuation—Case 8. Model: Qwen3-8B-Thinking. Includes the prompt, model response (excluding content within <think></think>), Gemini-2.5-Pro verification, and human evaluation result.}
\end{figure}

\begin{figure}
    \centering
    \includegraphics[width=\linewidth]{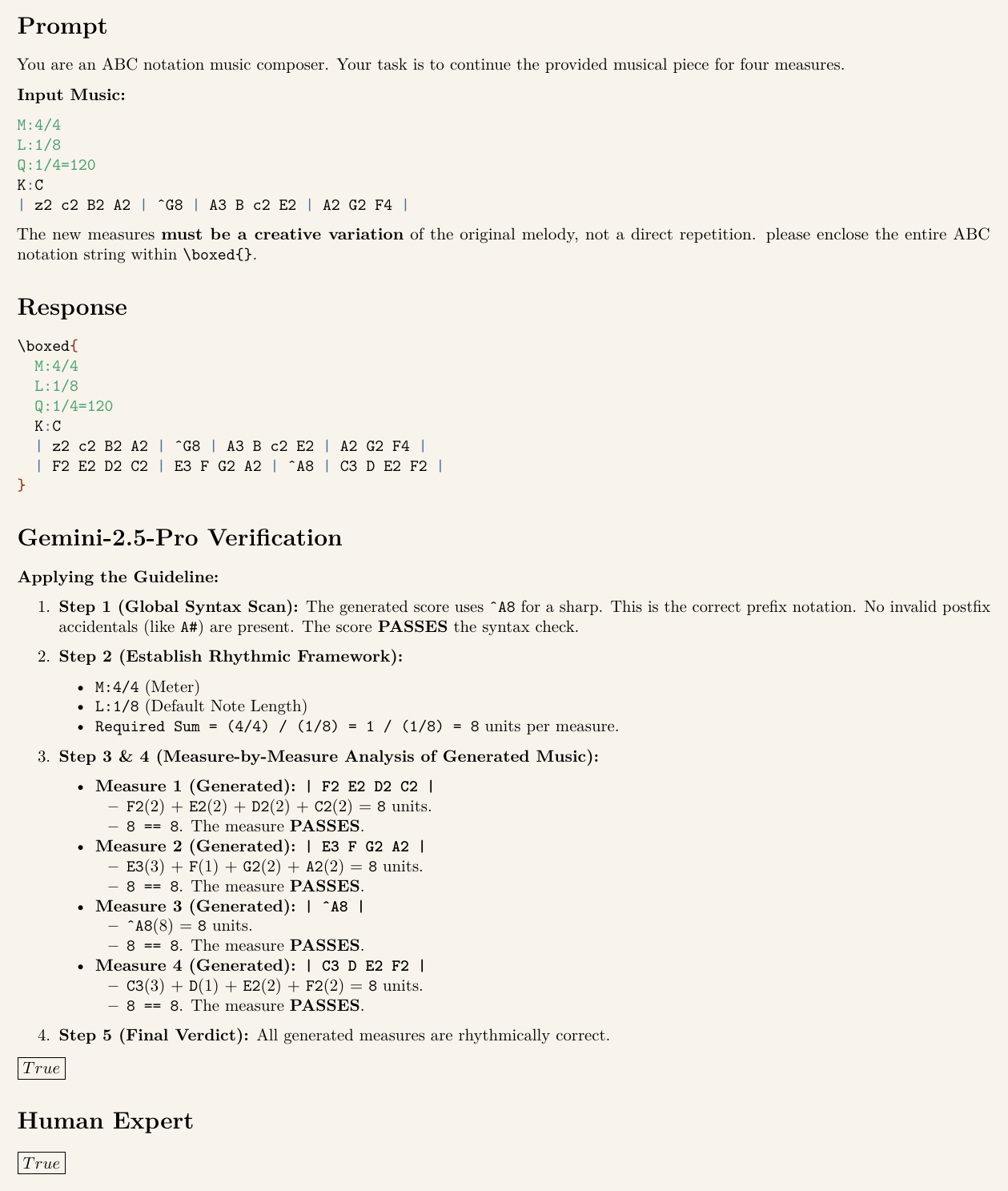}
    \caption{Sheet Music Continuation—Case 9. Model: Qwen3-8B-Thinking. Includes the prompt, model response (excluding content within <think></think>), Gemini-2.5-Pro verification, and human evaluation result.}
\end{figure}